\documentclass[10pt,a4paper]{IEEEtran}
\usepackage[utf8]{inputenc}
\usepackage{cite}
\usepackage{authblk}
\usepackage{amsfonts}
\usepackage{graphicx}
\usepackage{textcomp}
\usepackage{graphicx}
\usepackage{caption}
\usepackage{multirow}
\usepackage{booktabs}

\usepackage{color}
\usepackage{empheq}
\usepackage{algorithm}
\usepackage{algpseudocode}
\usepackage{subcaption}
\usepackage{hyperref}
\usepackage{url}
\usepackage{mathtools, cuted}
\usepackage{lipsum, color}
\usepackage[pdftex,dvipsnames]{xcolor}  

\usepackage{amsmath}
\usepackage{amssymb}
\usepackage{bm}
\usepackage{mathtools}
\usepackage{dsfont}
\usepackage{multirow}
\usepackage{leftidx}
\usepackage{soul}
\usepackage{textgreek}

\newcommand{\reviewerone}[1]{{\color{black}{#1}}}
\newcommand{\reviewertwo}[1]{{\color{black}{#1}}}
\newcommand{\reviewerthr}[1]{{\color{black}{#1}}}
\newcommand{\reviewerfur}[1]{{\color{black}{#1}}}
\def\BibTeX{{\rm B\kern-.05em{\sc i\kern-.025em b}\kern-.08em
    T\kern-.1667em\lower.7ex\hbox{E}\kern-.125emX}}

\begin{document}
\date{}
\renewcommand\footnotemark{}
\title{Learning a Model-Driven Variational Network for Deformable Image Registration}
\author{\centering Xi Jia, Alexander Thorley, Wei Chen, Huaqi Qiu, Linlin Shen, Iain B Styles, Hyung Jin Chang, \space\space\space \space \space \space \space \space \space \space \space \space \space \space \space \space \space \space \space \space  \space  Ales Leonardis, Antonio de Marvao, Declan P. O'Regan, Daniel Rueckert, \textit{Fellow, IEEE}, and Jinming Duan

\thanks{X. Jia, A. Thorley, W. Chen, I. Styles, HJ. Chang, A. Leonardis, and J. Duan are with the University of Birmingham, Birmingham, UK. I. Styles and A. Leonardis are Fellows of the Alan Turing Institute, London, UK. L. Shen is with the Shenzhen University, Guangdong, China. H. Qiu and D. Rueckert are with the Department of Computing, Imperial College London, London, UK. A. de Marvao and D. P. O'Regan are with the MRC London Institute of Medical Sciences, Imperial College London, London, UK. The correponding author is J. Duan (j.duan@cs.bham.ac.uk).}

\thanks{\textbf{This work has been submitted to the IEEE for possible publication. Copyright may be transferred without notice, after which this version may no longer be accessible.}}}  

\maketitle

\begin{abstract}

Data-driven deep learning approaches to image registration \reviewertwo{can be less accurate than conventional iterative approaches, especially when training data is limited}. To address this whilst retaining the fast inference speed of deep learning,  we propose VR-Net, a novel cascaded variational network for unsupervised deformable image registration. Using the variable splitting optimization scheme, we first convert the image registration problem, established in a generic variational framework, into two sub-problems, one with a point-wise, closed-form solution while the other one is a denoising problem. We then propose two neural layers (i.e. warping layer and intensity consistency layer) to model the analytical solution and a residual U-Net to formulate the denoising problem (i.e. \reviewerone{generalized denoising layer}). Finally, we cascade the warping layer, intensity consistency layer, and generalized denoising layer to form the VR-Net. Extensive experiments on three (two 2D and one 3D) cardiac magnetic resonance imaging datasets show that VR-Net outperforms state-of-the-art deep learning methods on registration accuracy, while maintains the fast inference speed of deep learning and the data-efficiency of variational model.

\end{abstract}

\section{Introduction}
\IEEEPARstart{I}{mage} registration maps a floating image to a reference image according to their spatial correspondence. The procedure typically involves two operations: 1) estimating the spatial transformation between the image pair; 2) deforming the floating image with the estimated transformation. In medical image analysis, registration is critical for many automatic analysis tasks such as multi-modality fusion, population modeling, and statistical atlas learning \cite{de2019DLIR,sotiras2013deformable}.

Image registration approaches can be broadly categorized into two major branches: intensity-based and landmark-based approaches. The intensity-based approaches can be either mono-modal or multi-modal. In mono-modal registration, a variational framework is often used in which the problem is framed as an optimization of the form: 
\begin{equation}
\label{eq:horn1981determining}
\min _{\boldsymbol{u}}\frac{1}{2}\int_{\Omega}|I_{1}(\boldsymbol{x}+\boldsymbol{u}(\boldsymbol{x}))-I_{0}(\boldsymbol{x})|^{2} \mathrm{d} \boldsymbol{x} + \lambda {\cal {R}}\left( \boldsymbol{u}(\boldsymbol{x}) \right), 
\end{equation}
where $I_0$ and $I_1$: $\left(\Omega \subseteq \mathbb{R}^{d}\right) \rightarrow \mathbb{R}$ represent the reference image and the floating image, respectively. $ \boldsymbol{u}(\boldsymbol{x})= ({u}_x(\boldsymbol{x}),{u}_y(\boldsymbol{x}))^ {\rm{T}}: \Omega \to {\mathbb {R}}^d$ denotes the deformation. \reviewerthr{In this paper, we study $d=2$ and $d=3$ which correspond to two-dimensional (2D) and three-dimensional (3D) cases.} The first term (i.e., data term) is the sum of squared differences, which is a \textit{similarity measure}. Minimization of the data term alone is typically an ill-posed problem with many possible solutions. Hence, the second regularization term is needed, which is normally chosen to control the smoothness of the deformation.

The variational model is among the most successful and accurate approaches to calculate a deformation between two images\cite{zach2007duality}. Given a specific regularization term, such a model has a clear mathematical structure and it is also well understood which mathematical space the solution lies in, e.g., Hilbert space \cite{fischer2002fast,beg2005computing,chen2019new}, bounded variation \cite{frohn2008multigrid,vishnevskiy2016isotropic}, etc. However, the variational model has limitations: (1) For each image pair, the hyper-parameter $\lambda$ needs to be tuned carefully to deliver a precise deformation. While a too small $\lambda$ leads to an irregular and non-smooth deformation, setting it too high reduces the deformation magnitude and therefore loses the ability to model large deformations. (2) The hand-crafted regularization term itself is another hyper-parameter, which is usually selected based on assumptions about the deformation. However, existing assumptions may be too simple to capture complex changes of image content associated with biological tissues. (3) The variational model is nonlinear and therefore needs to be optimized iteratively, which is very time-consuming especially for high-dimensional data inputs. 

Many deep learning approaches have been proposed for unsupervised deformable medical image registration \cite{balakrishnan2018unsupervised,zhang2018inverse,Zhao_2019_ICCV,qin2018joint,hering2019mlvirnet,krebs2019learning}. In order to learn a deformation, almost all of these learning-based approaches follow the formulation of $\boldsymbol{u} = f\left(I_0, I_1 | \textbf{W}\right)$, where $f$ is a convolutional neural network (CNN) and $ \textbf{W}$ denotes the weights of the CNN. These approaches are purely data-driven and differ from \reviewertwo{iterative variational} approaches in two main aspects. (1) 
\reviewertwo{Data-driven approaches take images as input and directly output the estimated deformations under a loss criterion, while traditional iterative approaches take an initial deformation as input, and output a final refined deformation which is built upon the previous deformations in the iterative optimization.} Whilst data-driven approaches often require substantial quantities of training data to reach an adequate level of performance, the iterative methods can work well in low data regimes. Additionally, the heavy data dependence of deep learning can result in a network that overfits the training data, and therefore lacks generalization abilities. 
(2) \reviewerthr{Classical iterative methods explicitly use prior and domain knowledge to construct a mathematical formulation. In contrast, data-driven methods implicitly learn prior and domain knowledge through the optimization of respective loss functions rather than explicitly building this knowledge into the network architecture itself.} Some researches \cite{schlemper2017deep,Hammernik_VN,duan2019vs,aggarwal2018modl} in image reconstruction have shown that integrating such knowledge into the network enables it learns better. Within image registration, data-driven approaches have not yet exceeded the accuracy of iterative approaches in some image registration tasks according to \cite{de2019DLIR,qiustacom19,Rueckert_Model_Data_Driven}, however, they have the advantage of significantly faster inference than their \reviewertwo{iterative optimization based} counterparts.

\begin{figure*}[ht] 
\centering  
{\includegraphics[width=0.85\textwidth]{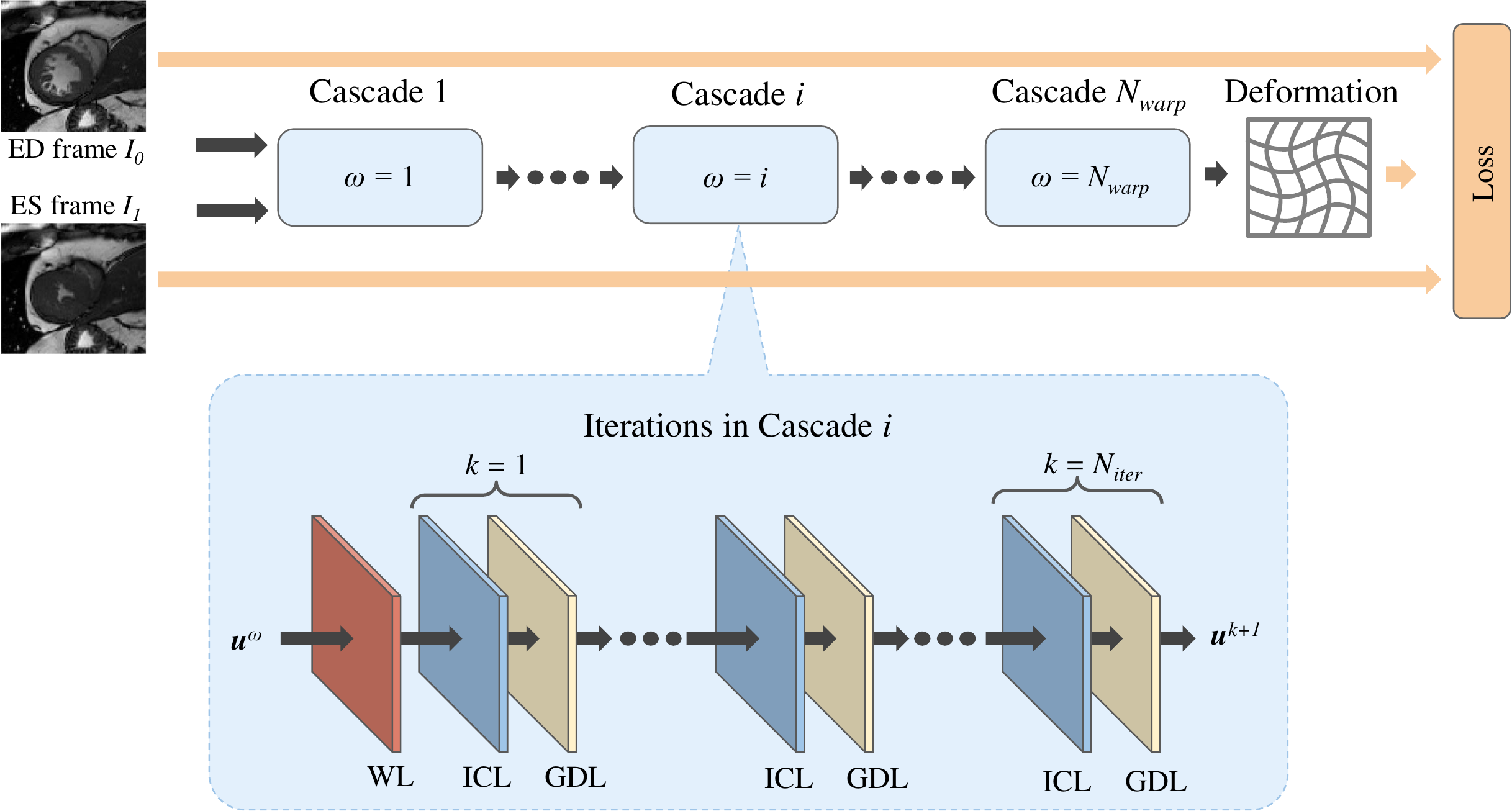}}

\caption{Overall architecture of the proposed VR-Net, where WL, ICL, and GDL denote the warping layer, intensity consistency layer, and generalized denoising layer, respectively. These layers are designed according to the minimization of a generic variational model for image registration. The number of cascades is controlled by $N_{warp} \times$ $N_{iter}$, which mimics the iterative process for the minimization.}
\label{fig:network}
\vspace{-10pt}
\end{figure*}

In order to take advantage of both methods, in this paper, we unify a data-driven and an iterative approach into one framework and propose a \reviewertwo{model-driven} variational registration network, which we term VR-Net as shown in Fig.~\ref{fig:network}. \reviewertwo{Note that we use the term iterative methods to denote traditional registration approaches such as TV-L$_1$ and FFD, which have a data term, a regularization term, and an optimization scheme to minimize the terms. We use the term data-driven to denote the recent deep learning methods that require large quantities of training data, following\cite{Rueckert_Model_Data_Driven}.
A model-driven approach is a learning-based approach that combines data-driven and iterative methods.
}

\reviewerthr{Specifically, with the help of a \textit{variable splitting} scheme for optimization, we decompose the original iterative variational problem into two sub-problems. One has a point-wise, closed-form solution and the other can be formulated as a denoising problem. Next, we formulate the point-wise, closed-form solution with a warping layer and an intensity consistency layer. We then propose a residual U-Net for the denoising sub-problem which can be regarded as a learnable regularization term embedded in the VR-Net, replacing the hand-crafted hyper-parameter seen in iterative variational methods. Finally, within the VR-Net we cascade the warping layer, the intensity consistency layer, and the generalized denoising layer to mimic the iterative process of solving a variational model.}

To evaluate the proposed VR-Net, we use two 2D publicly available cardiac MRI datasets, i.e., the UK Biobank dataset \cite{petersen2013imaging}, Automatic Cardiac Diagnosis Challenge (ACDC) dataset \cite{bernard2018deep} \reviewerthr{and one 3D cardiac MRI dataset (3D CMR)\cite{duan2019automatic}}. \reviewertwo{Extensive experiments on the datasets show that our VR-Net outperforms data-driven approaches with respect to registration accuracy and retains fast inference speed of deep networks.} Collectively, our main contributions are given as follows:
\begin{itemize}
\item We propose VR-Net for image registration. To our knowledge, this is the first model-driven deep learning approach tailored for this task. Such a network remedies the aforementioned limitations in both iterative methods and data-driven approaches and therefore paves a new way to solve the challenging task of image registration.
\item VR-Net embeds the mathematical structure from the minimization of a generic variational model into a neural network. The network mapping function $f$ therefore inherits prior knowledge from the variational model, whilst maintaining the data efficiency of the iterative methods and retaining the fast inference speed of data-driven registration methods. As such, it has the advantages from both communities and is shown to exceed state-of-the-art data-driven registration methods in terms of dice score and Hausdorf distance on both 2D and 3D datasets.
\item For iterative variational approaches to perform well for individual image pairs, one often needs to define a suitable regularization and then tune the corresponding regularization parameter $\lambda$. Instead, our VR-Net is trained with a global regularization parameter. This enables the model to effectively learn from the data the regularization term as well as the values of $\lambda$, which removes the need to tune on individual image pairs and thus resulting in a more generalizable model compared with its iterative counterparts.

\end{itemize}

\section{Related Works}
\textbf{Iterative Approaches:} Image registration using iterative approaches is performed for each image pair via iterative optimization of the transformation model parameters under both image intensity and regularization constraints. Affine transformations are typically firstly applied to handle global transformations such as rotation, translation, shearing and scaling. This is followed by a deformable transformation which has more degrees of freedom as well as higher capability to describe local deformations. There is a wide range of classical variational methods to account for local deformations such as diffusion models \cite{fischer2002fast}, total variation models \cite{frohn2008multigrid,vishnevskiy2016isotropic}, fluid models \cite{beg2005computing,chen2019new}, elastic models \cite{broit1981optimal,lin2012gene,vese2016variational}, biharmonic (linear curvature) models \cite{fischer2003curvature,modersitzki2004numerical}, mean curvature models \cite{chumchob2011fourth,chumchob2012improved}, optical flow models \cite{brox2004high,zach2007duality,wedel2009improved}, fractional-order variation models \cite{zhang2015variational,xu2019novel}, non-local graph models \cite{werlberger2010motion,ranftl2014non,papiez2016non}, etc. The free-form deformation (FFD) methods based on B-splines model \cite{rueckert1999nonrigid,lu2004fast} are able to accurately model global and local deformations with fewer degrees of freedom parameterized by control points. 

\textbf{Data-Driven Approaches:} Recently, researchers have started to shift their interests to unsupervised data-driven methods for medical image registration. These learning-based methods are normally trained with a large amount of paired images. By extending the spatial transformer network \cite{jaderberg2015spatial}, Balakrishnan et al. \cite{balakrishnan2018unsupervised,balakrishnan2019voxelmorph} proposed the VoxelMorph and evaluated the method on brain MRI image registration. Qin et al. \cite{qin2018joint} proposed a framework for joint registration and segmentation on cardiac MRI sequences, with their registration branch based on a Siamese-style, recurrent multi-scale network. de Vos et al. \cite{de2019DLIR} proposed a multi-stage, coarse-to-fine network (termed DLIR) for \reviewerfur{parametric registration. DLIR has two types of CNNs that account for global and local transformations, respectively. The global network estimates the affine transformation and the local networks predict the displacements parameterized by the B-spline control points.} \reviewerthr{The work \cite{GUO2020101769} proposed by Guo et al. is also a coarse-to-fine, multi-stage registration framework. However, this method estimates only rigid transformations while our method predicts dense displacements and performs nonrigid registration.} Zhao et al. \cite{Zhao_2019_ICCV} proposed a deep recursive cascade architecture, termed RC-Net. By cascading several base-nets, RC-Net achieved significant gains over VoxelMorph \cite{balakrishnan2018unsupervised} on both liver and brain registration tasks. Similar to RC-Net, the proposed VR-Net also uses a cascaded, end-to-end trainable network architecture. Within each cascade of VR-Net, however, we solve a point-wise, closed-form optimization problem induced by minimizing a generic variational model,  \reviewerthr{which is a major difference from other recursive \cite{Zhao_2019_ICCV} or multi-stage \cite{de2019DLIR,GUO2020101769} networks}. \\

\noindent \textbf{Model-Driven Approaches:} \reviewerfur{The authors in \cite{TPAMIChenTrainable,VariationalNetworks2017,Hammernik_VN} studied trainable variational networks to address supervised, linear image restoration and reconstruction problems, while we tackle an unsupervised, nonlinear image registration problem. Their methods are based on proximal gradient descent, and they learn the regularization term based on the Field of Experts (FoE)\cite{FOE1467533}. The nonlinearity (derivative of the potential function in the regularization) in their method is imposed by the radial basis kernels. The optimization of their method is done through the specialized inertial incremental proximal method (IIPG), which is not implemented in the standard deep learning framework (e.g. Pytorch) and therefore may be difficult to generalize to other problems. In contrast, our network is based on a linearized variable splitting method, one advantage of which is that we can impose the exact data term in each cascade which cannot be done by gradient-based methods. Our regularization is formulated as a CNN, where the nonlinearity is imposed by the activation functions (such as ReLU) and the parameters are optimized by Adam in a standard deep learning framework. There also exist works \cite{fan2018end, zhang2017learning, schlemper2017deep,aggarwal2018modl,duan2019vs} that have explored variational formulations in the deep learning framework. However, instead of image registration, they were used either for image restoration and reconstruction or for video understanding. Recently, \reviewerfur{Blendowski et al. \cite{BLENDOWSKI2021101822} proposed a supervised iterative descent algorithm (SUITS) for multi-modal image registration, which is similar to our work. SUITS uses a CNN to extract image features, which are then plugged into the Horn and Schunck (HS) model \cite{horn1981determining} to compute displacements. In other words, they need to solve an iterative model within the network each time when new displacements are required. This method can be expensive because (1) the HS model needs to have many data terms (12 in their paper) in order to align all extracted features; (2) solving the HS model itself is costly and requires iterations; and (3) they need to solve the HS model many times within the network. In contrast, we do not need to iteratively solve any optimization model within our network. Instead, we use the iterative process for optimization only to guide the design of network architecture. Moreover, unlike Blendowski et al's work which uses an algebraic multigrid solver (AMG) to solve the linear system of equations, all subproblems (network layers) in our method have closed-form, point-wise solutions.}}

\section{Generic Variational Method}
In this section, we study a more general variational model for image registration, which is given by
\begin{equation} \label{eq:variational}
\min _{\boldsymbol{u}}\frac{1}{s}\int_{\Omega}|I_{1}(\boldsymbol{x}+\boldsymbol{u}(\boldsymbol{x}))-I_{0}(\boldsymbol{x})|^{s} \mathrm{d} \boldsymbol{x} + \lambda {\cal {R}}\left( \boldsymbol{u}(\boldsymbol{x}) \right) , 
\end{equation}
where the variables in this formulation have the same meaning as in Eq.~\eqref{eq:horn1981determining}. The objective is to find the optimal deformation $\boldsymbol{u}^{*}(\boldsymbol{x}):(\Omega \subseteq {\mathbb {R}}^d) \to {\mathbb {R}}^d$, that minimizes the formulation. Within the data term, $s=1$ corresponds to $L_1$ estimation that is robust to outliers, while $s=2$ gives the estimation based on the sum of squared difference. The second term is a generic regularization term, which imposes a smoothness constraint on the deformation. The hyper-parameter $\lambda$ controls the smoothness of the solution. However, for different applications it is non-trivial to select a $\lambda$ and a regularization term that are both suitable. 

In the data term, we notice that the non-linearity in the function $I_1({\boldsymbol{x} + \boldsymbol{u}})$ with respect to $\boldsymbol{u}$ poses a challenge to optimize Eq.~\eqref{eq:variational}. To benefit from closed-form solutions, we use the Gauss–Newton algorithm to handle Eq. \eqref{eq:variational} \reviewertwo{\cite{zach2007duality,baker2004lucas}}. By employing the first-order Taylor expansion at $\boldsymbol{u}^\omega$, we end up with solving the following alternative problem:
\begin{subequations}
  \begin{equation}
    {I_1}\left( {\boldsymbol{x} + \boldsymbol{u}} \right) =  {I_1}\left( {\boldsymbol{x} + \boldsymbol{u}^\omega} \right) +  \langle \nabla I_1({\boldsymbol{x} + \boldsymbol{u}^\omega}),\boldsymbol{u} - \boldsymbol{u}^\omega\rangle  \vspace{5pt} \label{eq:taylor}
  \end{equation}
  \begin{equation}
    \boldsymbol{u}^{\omega+1} = {\arg\min}_{\boldsymbol{u}}\frac{1}{s}\int_{\Omega}|\rho (\boldsymbol{u}) |^{s} d\boldsymbol{x} + \lambda {\cal {R}}\left( \boldsymbol{u} \label{eq:linearminpro} \right),
  \end{equation}
\end{subequations}
where 
\begin{equation} \label{eq:rho}
\rho (\boldsymbol{u}) = {I_1}\left( {\boldsymbol{x} + \boldsymbol{u}^\omega} \right) + \langle \nabla I_1({\boldsymbol{x} + \boldsymbol{u}^\omega}),\boldsymbol{u} - \boldsymbol{u}^\omega\rangle - I_0(\boldsymbol{x}).
\end{equation}
In Eq. \eqref{eq:taylor}, $\nabla$ is the gradient operator, $\nabla I_1$ represents partial derivatives of $I_1$, $\langle \cdot, \cdot \rangle$ denotes the inner product and ${\omega}$ denotes the ${\omega}^{th}$  iteration. The linearized version of Eq. \eqref{eq:variational}, seen in Eq. \eqref{eq:linearminpro}, must to be solved iteratively. As the data term in Eq. \eqref{eq:linearminpro} is in a linear, convex form, one can derive a closed-form solution. Of note, to solve Eq. \eqref{eq:variational} approximately, one needs to iterate between Eq. \eqref{eq:taylor} and Eq. \eqref{eq:linearminpro}, meaning that there exist two loops in the resulting numerical implementation. 

The regularization ${\cal {R}}\left( \boldsymbol{u} \right)$ has many choices depending on what the final deformation $\boldsymbol{u}^*$ looks like, such as piecewise smooth, piecewise constant, etc. A widely used choice for it is the Total Variation (TV), which is a powerful regularization that allows discontinuities in the resulting deformation. However, a major challenge for those hand-crafted regularizations is that they may not be optimal for more complex, task-specific applications. To circumvent these, we propose an end-to-end, trainable VR-Net detailed in Section \ref{sec:architecture}.

\subsection{Variable Splitting}
\label{sec:VS}
To design an appropriate VR-Net, we first adopt a variable splitting scheme to minimize the linearized variational model Eq. \eqref{eq:linearminpro}. Specifically, we introduce an auxiliary splitting variable $\boldsymbol{v}: (\Omega \subseteq \mathbb {R}^d) \to {\mathbb {R}^d}$, converting  Eq. \eqref{eq:linearminpro} into the equivalent constrained minimization problem
\begin{equation} \nonumber
\min _{\boldsymbol{u}, \boldsymbol{v}} \frac{1}{s} \int_\Omega | \rho (\boldsymbol{u}) |^s d \boldsymbol{x} + \lambda {\cal R}(\boldsymbol{v}) \quad s.t. \quad \boldsymbol{u} = \boldsymbol{v}.
\end{equation}
The introduction of the constraint $\boldsymbol{u} = \boldsymbol{v}$ above decouples $\boldsymbol{u}$ in the regularization term from the data term, therefore a multi-channel denoising problem can be explicitly constructed and a closed-form, point-wise solution can be derived. Using the penalty function method, we then add the constraint back into the model and minimize the single problem
\begin{equation} \nonumber
\min _{\boldsymbol{u}, \boldsymbol{v}} \frac{1}{s} \int_\Omega | \rho (\boldsymbol{u}) |^s d \boldsymbol{x} + \lambda {\cal R}(\boldsymbol{v}) + \frac{\theta}{2} \int_\Omega |\boldsymbol{v} - \boldsymbol{u}|^2d\boldsymbol{x},
\end{equation}
where $\theta$ is the introduced penalty weight. To solve the multi-variable minimization problem, one needs to minimize it with respect to $\boldsymbol{u}$ and $\boldsymbol{v}$ separately.

1) $\boldsymbol{u}$-\textit{\textbf{subproblem}} is a linear problem and handled by considering the following minimization problem
\begin{equation} \nonumber
 \boldsymbol{u}^{k+1} = \mathop {{\rm{argmin}}}\limits_{\boldsymbol{u}} \frac{1}{s} \int_\Omega | \rho (\boldsymbol{u}) |^s d \boldsymbol{x} + \frac{\theta}{2} \int_\Omega |\boldsymbol{v}^k - \boldsymbol{u}|^2d\boldsymbol{x}, 
\end{equation}
the solution of which depends on the order of $s$. In the case of $s=1$, the solution is given by the following thresholding equation
\begin{equation}\label{eq:soft1}
\boldsymbol{u}^{k+1} = \boldsymbol{v}^{k}  - \frac{{{{\hat{z} }}}}{{\max \left( {\left| {{{\hat{z} }}} \right|,1} \right)}} \frac{\nabla I_1}{\theta},
\end{equation}
where $\hat{z} = \theta{\rho({ \boldsymbol{v}^{k}  })}/{(|\nabla I_1|^2+ \epsilon)}$ and $\epsilon$ is a small positive value added to avoid division by zero to prevent vanishing gradients in the image. \reviewerfur{In \hyperlink{A1}{Appendix 1} we develop a novel primal-dual method to derive this solution (\ref{eq:soft1}). Our new derivation allows the proposed method to easily adapt to vector images which usually appearing in data terms that use image patch or (higher-order) gradient information \cite{vogel2013evaluation,papenberg2006highly}}.


\reviewerfur{In the case of $s=2$, the respective problem is differentiable and we can derive the following Sherman–Morrison formula by directly differentiating this subproblem with respect to $\boldsymbol{u}$
\begin{equation} \label{eq:ShermanMorrison}
    ({\bf{J}} {\bf{J}} ^{\rm{T}} + \theta {\mathds{1}}) (\boldsymbol{u}- \boldsymbol{u}^\omega) = \theta ({\boldsymbol{v}}^k - \boldsymbol{u}^\omega )   - {\bf{J}}(I_1 - I_0),
\end{equation}
where ${\bf{J}} {\bf{J}} ^ {\rm{T}}$ (where ${\bf{J}}=\nabla I_1$) is the outer product and ${\mathds{1}}$ is an identity matrix. Due to the identity matrix, the Sherman–Morrison formula will lead a close-form, point-wise solution to $\boldsymbol{u}^{k+1}$. In \hyperlink{A2}{Appendix 2}, we show the detailed derivations in both 2D and 3D.} 

2) $\boldsymbol{v}$-\textit{\textbf{subproblem}} is handled by considering the following minimization problem
\begin{equation} \label{eq:denoising}
\boldsymbol{v}^{k+1} = \mathop {{\rm{argmin}}}\limits_{\boldsymbol{v}} \lambda {\cal R}(\boldsymbol{v}) + \frac{\theta}{2} \int_\Omega |\boldsymbol{v} - \boldsymbol{u}^{k+1}|^2d\boldsymbol{x}.
\end{equation}
Given a known $\boldsymbol{u}^{k+1}$, this problem essentially is a denoising problem with the generic regularization ${\cal R}(\boldsymbol{v})$. Note that we assume the noise here is additive and follows a Gaussian distribution. For example, if the regularization ${\cal R}(\boldsymbol{v})$ is a total variation (TV), then it is a TV denoising problem, as in Zach's paper \cite{zach2007duality}.

Putting these derivations together, we have Algorithm 1 to minimize Eq. \eqref{eq:variational} using the variable splitting. Since Taylor expansion is used to linearize the non-linear function, Eq.~\eqref{eq:taylor} holds only if the resulting deformation $\boldsymbol{u}^*$ is small. As such, we adopt an extra $warping$ operation (via $\boldsymbol{u}^\omega$) in Algorithm 1, i.e., $I_1^\omega = I_1(\boldsymbol{x}+\boldsymbol{u}^\omega)$. With $warping$, we can break down a large deformation into $N_{warp}$ small ones, each of which can be solved iteratively and optimally. The total iterations for the algorithm is $N_{warp} \times N_{iter}$.

\begin{algorithm}
  \caption{VS for generic variational registration model}\label{euclid}
  \begin{algorithmic}[1]
      \State \textbf{Inputs} : $I_0$, $I_1$ and $(\theta,\lambda, N_{warp}, N_{iter})$.  
      \State \textbf{Initialize} : $\boldsymbol{u}^1$ and ${\boldsymbol{v}^1}$.
          \For {$\omega =1 : N_{warp}$} \Comment{\# Taylor expansions}
              \State $I_1^\omega$ = ${warping}$($I_1$, $\boldsymbol{u}^{\omega}$)
              \While {$k < N_{iter}$} \Comment{\# iterations}
                  \State update $\boldsymbol{u}^{k+1}$ via \eqref{eq:soft1}, \eqref{eq:s1} or \eqref{eq:s2} with $I_1=I_1^\omega$\; 
                  \State $\boldsymbol{v}^{k+1}=denoiser(\boldsymbol{u}^{k+1})$
              \EndWhile\label{euclidendwhile}
              \State $\boldsymbol{u}^\omega$ = $\boldsymbol{u}^{k+1}$ \; 
          \EndFor\label{euclidendfor}
      \State \Return $\boldsymbol{u}^*=\boldsymbol{u}^\omega$  \Comment{\# return final solution}
  \end{algorithmic}
\end{algorithm}

\section{Learning a Variational Registration Network}
So far, we have shown how the variable splitting scheme can be derived to tackle the generic variational registration model. We first handle the original problem Eq. \eqref{eq:variational} with the Gauss-Newton method. For the resulting linearized minimization problem Eq. \eqref{eq:linearminpro} we have two sub-problems, one with a closed-form, point-wise solution for either choice of the $s$ and one a denoising problem with ${\cal R} {(\boldsymbol{u})}$. As of yet, we have not defined the exact form of $denoiser$ in Algorithm 1. In the following section, we will detail the full VR-Net architecture, and show how a residual CNN is used as our $denoiser$ to solve the second denoising sub-problem.

\begin{figure*}[ht] 
\centering  
{\includegraphics[width=0.85\textwidth]{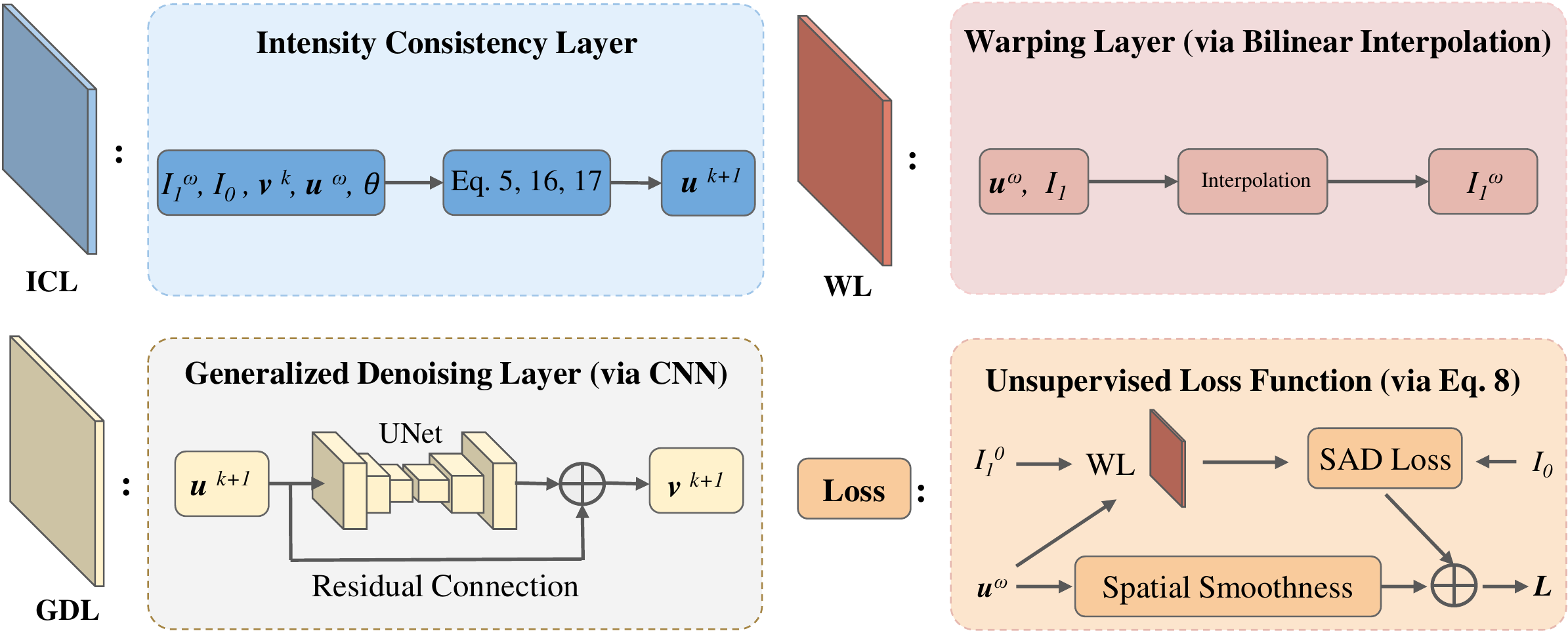}}
\caption{Detailed structure of each layer in VR-Net. WL, ICL, and GDL stand for warping layer, intensity consistency layer, and generalized denoising layer, respectively.}
\label{fig:blocks}
\end{figure*}

\subsection{Network Architecture}
\label{sec:architecture}
We construct the proposed VR-Net by unrolling the iterative procedure in Algorithm 1. Fig.~\ref{fig:network} depicts the resulting network architecture. There are two types of cascade in the architecture to learn a large displacement: (1) $cascade$-$iter$ indicated by $k\in \{1,...,N_{iter}\}$, stands for the inner loop in Algorithm 1; (2) $cascade$-$warp$ indicated by $\omega \in \{1,...,N_{warp}\}$, corresponds to the outer loop in Algorithm 1. Note that $cascade$-$warp$ contains multiple nested $cascade$-$iter$s. In Fig.~\ref{fig:blocks}, we show the three computational layers contained in the network, which are the \textit{warping layer} (WL), the \textit{intensity consistency layer} (ICL) and the \reviewerone{\textit{generalized denoising layer} (GDL)}. They respectively correspond to Step 4, 6 and 7 in Algorithm 1. 

\textbf{\textit{Warping layer}} is achieved by using a bilinear interpolation for 2D images, \reviewerfur{following the spatial transformer networks \cite{jaderberg2015spatial}}. Recall that the $warping$ operation is defined in Algorithm 1 by $I_1^\omega = I_1(\boldsymbol{x}+\boldsymbol{u}^\omega)$, 
where $\boldsymbol{u}^\omega$ is the estimated displacement. The bilinear interpolation is continuous and piecewise smooth, and the partial gradients with respect to $\boldsymbol{u}^\omega$ can be derived\reviewerfur{ as in \cite{jaderberg2015spatial}. The 2D warping layer can be easily extended to 3D to transform 3D volumes, similarly in \cite{balakrishnan2019voxelmorph}}. In Fig.~\ref{fig:blocks}, we show the computational graph of this layer, which takes $\boldsymbol{u}^\omega$ and $I_1$ as the inputs and outputs the warped image $I_1^\omega$.

\textbf{\textit{Intensity consistency layer}} is crucial as it effectively imposes intensity consistency between the warped image ($I_1^w$) and the target image ($I_0$) such that the data term in Eq. \eqref{eq:variational} can be minimized. Fig.~\ref{fig:blocks} presents the computational graph of this layer. Specifically, the input $I_1^w$ from the upstream warping layer, concurrently with $I_0$, $\boldsymbol{v}^{k}$, $\boldsymbol{u}^{\omega}$ and $\theta$, are passed through Eq. \eqref{eq:soft1}, \eqref{eq:s1} or \eqref{eq:s2} to produce $\boldsymbol{u}^{k+1}$, which then feeds the downstream generalized denoising layer. Note that the calculations in this layer are both computationally efficient and numerically accurate thanks to the existance of point-wise, analytical solutions from Eq. \eqref{eq:soft1}, \eqref{eq:s1} or \eqref{eq:s2}. The penalty weight $\theta$ is often manually selected in iterative methods, however in this paper we instead make it a learnable parameter. 

\textbf{\textit{Generalized denoising layer}} is a residual U-Net that explicitly defines \textit{denoiser} in Algorithm 1. As illustrated in Fig.~\ref{fig:blocks}, we intend to denoise a two-channel displacement $\boldsymbol{u}^{k+1}$ with the residual U-Net and produce its denoised version $\boldsymbol{v}^{k+1}$ for ICL in next iteration. Since the input and output of ICL and GDL are both deformations, it is natural that we can adopt a residual connection between two adjacent cascades. As the generalized denoising layer represents the denoising subproblem Eq. \eqref{eq:denoising}, it implicitly absorbs the hyper-parameters $\lambda$ and $\theta$ and thus there is no need to tune them manually. Note that while we use a residual U-Net as the backbone here, our setup is generic and therefore allows for the incorporation of more advanced denoising CNN architectures. 

The function in Eq.~\eqref{eq:soft1} needs special attention when implemented as a neural layer. Although it is a continuous and piecewise smooth function, it is non-differentiable. As such, the concept of sub-gradients must be used during network back-propagation. As a result, this gives us a sub-differentiable mechanism with respect to network parameters, which allows loss gradients to flow back not only to the GDL and WL but also to the ICL.

\subsection{Network Loss and Parameterizations}
\label{sec:para}
\textbf{Network loss}: While the design of VR-Net architecture follows the philosophy of conventional optimization for iterative methods, training the network parameters is another optimization process, for which a loss function must be explicitly formulated. Due to the absence of ground truth transformation in medical imaging, we adopt an unsupervised loss function, using the floating image $I_1$, the reference image $I_0$ and the predicted deformation $\boldsymbol{u}$. The loss has the form of
\begin{equation}\label{eq:loss}
    \begin{split}
    {\cal L}({\bf{\Theta}})  = \min_{\bf{\Theta}}  \frac{1}{N} & \sum\limits_{i=1}^N \| I_1^i(\boldsymbol{x}+ {\boldsymbol{u}_i}({\bf{\Theta}}))  - I_0^i(\boldsymbol{x}) \|_1   \\ & + \frac{\alpha}{N} \sum\limits_{i=1}^N \|\nabla \boldsymbol{u}_i({\bf{\Theta}}) \|_2^2,  
    \end{split}
\end{equation}
where $N$ is the number of training image pairs, ${\bf{\Theta}}$ are the network parameters to be learned and $\alpha$ is a hyper-parameter balancing the two losses. Note that the first loss defines the similarity between the warped images and the reference images and the second loss defines the smoothness on the resulting displacements. The graph representation of the two loss functions is detailed in Fig.~\ref{fig:blocks}.

\reviewerone{Despite the model-driven components of our VR-Net, the method is essentially a deep learning approach so it also requires a smoothness parameter $\alpha$ that regularizes the learned displacements for the whole dataset. In contrast to the manual tuning of $\theta$ in Eq. \eqref{eq:soft1} and \eqref{eq:ShermanMorrison} which is required for each test pair image in traditional iterative methods, this smoothness parameter $\alpha$ is only tuned in the training set and used for inference without further optimization.} Note that $\alpha$ is not necessary if we instead use the MSE loss between predicted and ground-truth deformations as our loss function. \reviewerone{As shown in the Fig. \ref{fig:network}, we do not evaluate the loss function \eqref{eq:loss} at every cascade and only use it once at the very end of the VR-Net.}

\textbf{Parameterizations}: The network learnable parameters ${\bf{\Theta}}$ include both the residual U-Net parameters ${\bf{W}}$ in GDL layers and the penalty weights $\theta$ in the ICLs. Recall that in VR-Net (see Fig.~\ref{fig:network}) we have $cascade$-$iter$ and $cascade$-$warp$, and therefore each GDL and ICL layer has a set of parameters ${\bf{W}}$ and $\theta$, respectively. We experimented with two parameterization settings: ${\bf{\Theta}^1} = \left\{\bf{W}, \theta \right\}$ and ${\bf{\Theta}^2} = \{ \{ {\bf{W}}_{k,\omega}, \theta_{k,\omega} \}_{k = 1}^{N_{iter}}\}_{\omega=1}^{N_{warp}}$. For ${\bf{\Theta^1}}$, we let the parameters $\bf{W}$ and $\theta$ respectively be shared by the DLs and the ICLs across $cascade$-$warp$ and $cascade$-$iter$. In ${\bf{\Theta^2}}$, the parameters are not shared in either $cascade$-$iter$ or $cascade$-$warp$, meaning that each layer (GDL or ICL) has its own learnable parameter. For both parameterizations we experimented with, backpropagation is employed to minimize the loss with respect to the network parameters ${\bf{\Theta}}$ in an end-to-end fashion.

\subsection{Initialization}
\reviewerfur{While data-driven methods take image pairs as input and directly output the estimated deformations, we need an initial displacement as input of VR-Net as stated in Step 2 of Algorithm 1. The initial displacement is then refined by the iterative process. In this paper, we proposed 3 different initialization strategies.}
\reviewertwo{The first strategy is to initialize the $\boldsymbol{u}^1$ and ${\boldsymbol{v}^1}$ with zeros, which is used in the original TV-$\rm{L}_1$ paper \cite{zach2007duality}. The second strategy is using the Gaussian noise as initialization. However, initializing  $\boldsymbol{u}^1$ and ${\boldsymbol{v}^1}$ with zeros or noise is not necessarily the optimal choice. Inspired by \cite{fan2018end}, we propose to learn the initialization from the data by concatenating a U-Net prior to the first WL.}
\reviewerfur{Note that the additional concatenated U-Net is not pre-trained. It is a part of the VR-Net and its weights are updated along with the whole VR-Net during the training process.}

We evaluate the three different initialization strategies in \ref{sec:initialResult} and show that the registration performance benefits from making the initialization learnable.
\vspace{-5pt}
\section{Experimental results}
In this section, we introduce the datasets and quantitative metrics used for experiments. Then we describe the implementation details of the proposed method as well as ablation studies using different configurations. Finally, we compare the proposed VR-Net with state-of-the-art methods, including both iterative methods and data-driven approaches. 

\subsection{Datasets and Quantitative Metrics} 
\textbf{2D Datasets}: We evaluate the proposed VR-Net on the UK Biobank dataset \cite{petersen2013imaging} and the ACDC dataset \cite{bernard2018deep}. The UK Biobank \cite{petersen2013imaging} is a large scale cardiac MRI image dataset designed for cohort studies on 100,000 subjects. MRI scans in this dataset were acquired from healthy volunteers by using the same equipment and protocols, and the in-plane and through-plane resolutions are $1.8mm$ and $10mm$, respectively. We randomly select 220 subjects and split them into 100, 20, and 100 for training, validation, and testing, respectively. The ACDC dataset \cite{bernard2018deep} was created from real clinical exams. Acquisitions were obtained over a 6 year period with two MRI scanners of different magnetic strengths. The dataset is composed of 150 patients evenly divided into 5 types of pathology. We select the 100 subjects that have ground truth segmentation masks for experiments. We split these subjects into 40, 10, and 50 for training, validation, and testing, respectively. Since the in-plane resolution varies from 1.34 to 1.68$mm$, we resample all the images to $1.8mm$ before experiments. For both datasets, we perform experiments on only basal, mid-ventricular, and apical image slices.

\reviewerthr{\textbf{3D Dataset}: The 3D CMR dataset \cite{duan2019automatic} used in our experiments consists of 220 pairs of 3D high-resolution (HR) cardiac MRI images corresponding to the end diastolic (ED) and end systolic (ES) frames of the cardiac cycle. HR imaging requires only one single breath-hold and therefore introduces no inter-slice shift artifacts. All images are resampled to 1.2$\times$1.2$\times$1.2$mm^3$ resolution and cropped or padded to matrix size 128$\times$128$\times$96. To train comparative deep learning methods and tune hyperparameters in different methods, the dataset is split into 100/20/100 corresponding to training, validation, and test sets. We report final quantitative results on the test set only.}

Due to the absence of ground truth deformations for these datasets, we evaluate the performance of different methods using the segmentation masks of left ventricle cavity (LV), left ventricle myocardium (Myo), and right ventricle cavity (RV). Specifically, we calculate the deformation between ES and ED frames and then warp the ES segmentation using the deformation. Based on the warped ES segmentation and the ground truth ED segmentation, we compute Dice score and Hausdorff distance (HD) score \cite{bai2018automated}. The Dice score varies from $0$ to $1$, with higher values corresponding to a better match. The HD is measured on the outer contour of each anatomical structure: LV, Myo, and RV. It is on an open-ended scale, with smaller values implying a better result.

\subsection{Implementation Details}
\label{section:implementation}
We implement the proposed 2D VR-Net with U-Net \cite{ronneberger2015u} as the backbone for all \reviewerone{generalized denoising layers}. \reviewertwo{We used the original U-Net architecture in \cite{ronneberger2015u} and no further optimization of the architecture is performed.} As the input and output of such layers are displacements, we also apply a residual connection to the U-Net. To numerically discretize the partial derivatives $\nabla I_1$ in Eq. \eqref{eq:soft1} and Eq. \eqref{eq:ShermanMorrison} and $\nabla \boldsymbol{u}$ in the loss Eq. \eqref{eq:loss}, the central finite difference method is adopted. To train the 2D VR-Net, the batch size is set to 10 pairs of images.  $\alpha$ in Eq. \eqref{eq:loss} is selected using the \reviewertwo{grid-search} strategy on the validation set and set 0.1 for UK Biobank and 0.05 for ACDC. For training, we use the basal, mid-ventricular, and apical image slices in all frames from all subjects in the training set. During inference, we evaluate the 2D VR-Net and other comparative approaches using the three slices at the ED and ES phases from all subjects in the test set. This is because we only have manual segmentation masks at the two phases. \reviewerthr{Extending the 2D VR-Net to 3D is straightforward. The major difference between the 2D and 3D VR-Net is the generalized denoising layer, for 3D, we adopt a lighter 5-level hierarchical U-shape network from \cite{Mok_2020_CVPR} as the backbone. The training batch size of 3D VR-Net is set to 2. $\alpha$ in Eq. \eqref{eq:loss} is also selected using the validation set and set to be 0.0001 for the 3D CMR dataset.}

Both 2D and 3D VR-Nets are implemented with Pytorch \cite{paszke2019pytorch} and trained using a GeForce 1080 Ti GPU with 11GB RAM. An Adam optimizer \cite{kingma2014adam} with two beta values of 0.9 and 0.999 is used and the initial learning rate is set 0.0001. Note that we train our VR-Net using each dataset separately. For UK Biobank, the maximum iterations are 50,000 and the learning rate is gradually reduced after 25,000 iterations. For ACDC, the maximum iterations are 20,000 and the learning rate is gradually reduced after 10,000 iterations. For the 3D CMR dataset, the maximum iterations are 30,000 and the learning rate keeps fixed during training.  \reviewerfur{Due to the limitation of GPU memory, the maximum cascade number we could afford is 6 and 2 for 2D and 3D VR-Net, respectively. VR-Net is memory intensive as it has multiple cascaded GDL, however, it is very efficient in terms of speed during inference, as listed in Section \ref{sec:sota}. Note the memory dependencies can be reduced by using lighter CNN architectures in GDL.}

\begin{table*}[t]
\vspace{-10pt}
\centering
\caption{Comparison of image registration performance on two datasets using different configurations for the proposed VR-Net. Dice (HD) score is computed by averaging that of LV, Myo and RV at the basal, mid-ventricular and apical image slices from all subjects in the test set. Mean and standard deviation (in parenthesis) are reported.}
\resizebox{0.99\textwidth}{!}{
\begin{tabular}{cccccccccc}
\hline
 & \multicolumn{2}{c}{UK Biobank} & \multicolumn{2}{c}{ACDC} &  & \multicolumn{2}{c}{UK Biobank} & \multicolumn{2}{c}{ACDC} \\ \cline{2-5} \cline{7-10} 

 \multirow{-2}{*}{Methods} & Dice & HD & Dice & HD &  \multirow{-2}{*}{Methods} & Dice & HD & Dice & HD \\ \hline
 R-$\rm{L}_2$-1$\times$1 & .785(.047) & 10.69(3.05) & .860(.058) & 6.74(2.40) &  U-$\rm{L}_2$-1$\times$1 & .784(.046) & 10.65(3.03) & .860(.062) & 6.72(2.53) \\
 R-$\rm{L}_2$-2$\times$1 & .795(.046) & 10.69(3.15) & .850(.063) & 6.67(2.19) &  U-$\rm{L}_2$-2$\times$1 & .791(.045) & 10.52(3.06) & .861(.058) & 6.57(2.43) \\
 R-$\rm{L}_2$-2$\times$2 & .798(.046) & 10.49(3.12) & .867(.054) & 6.60(2.38) &  U-$\rm{L}_2$-2$\times$2 & .793(.044) & 10.48(3.09) & \textbf{.866(.056)} & 6.58(2.53) \\
 R-$\rm{L}_2$-3$\times$2 & .799(.044) & 10.63(3.18) & \textbf{.872(.052)} & \textbf{6.44(2.38)} &  U-$\rm{L}_2$-3$\times$2 & .802(.043) & \textbf{10.32(3.09)} & .866(.060) & \textbf{6.52(2.43)} \\
 R-$\rm{L}_2$-6$\times$1 & \textbf{.804(.043)} & \textbf{10.26(3.07)} & .869(.054) & 6.65(2.50) &  U-$\rm{L}_2$-6$\times$1 & \textbf{.803(.043)} & 10.47(3.14) & .856(.062) & 6.76(2.52) \\ \hline
 R-$\rm{L}_1$-1$\times$1 & .779(.048) & 10.77(3.03) & .853(.061) & 6.75(2.53) &  U-$\rm{L}_1$-1$\times$1 & .781(.047) & 10.77(3.07) & .854(.063) & 6.88(2.58) \\
 R-$\rm{L}_1$-2$\times$1 & .789(.047) & 10.62(3.13) & .865(.058) & 6.51(2.42) &  U-$\rm{L}_1$-2$\times$1 & .783(.048) & 10.74(3.04) & .858(.062) & 6.77(2.54) \\
 R-$\rm{L}_1$-2$\times$2 & .794(.045) & 10.58(3.00) & .865(.060) & 6.48(2.38) &  U-$\rm{L}_1$-2$\times$2 & .793(.046) & 10.56(3.01) & .867(.058) & 6.55(2.48) \\
 R-$\rm{L}_1$-3$\times$2 & .796(.046) & 10.54(3.16) & \textbf{.873(.050)} & \textbf{6.33(2.13)} &  U-$\rm{L}_1$-3$\times$2 & \textbf{.797(.045)} & \textbf{10.53(3.14)} & \textbf{.872(.052)} & \textbf{6.39(2.50)} \\
 R-$\rm{L}_1$-6$\times$1 & \textbf{.800(.045)} & \textbf{10.35(3.10)} & .850(.063) & 6.67(2.19) &  U-$\rm{L}_1$-6$\times$1 & .790(.098) & 10.61(3.10) & .845(.068) & 7.03(2.62) \\ \hline
\end{tabular}
}
\label{tab:cascade}
\end{table*}

\begin{figure*}[ht]
\begin{subfigure}{0.33\textwidth}
  \centering
  \includegraphics[width=0.99\linewidth,height=0.81\textwidth]{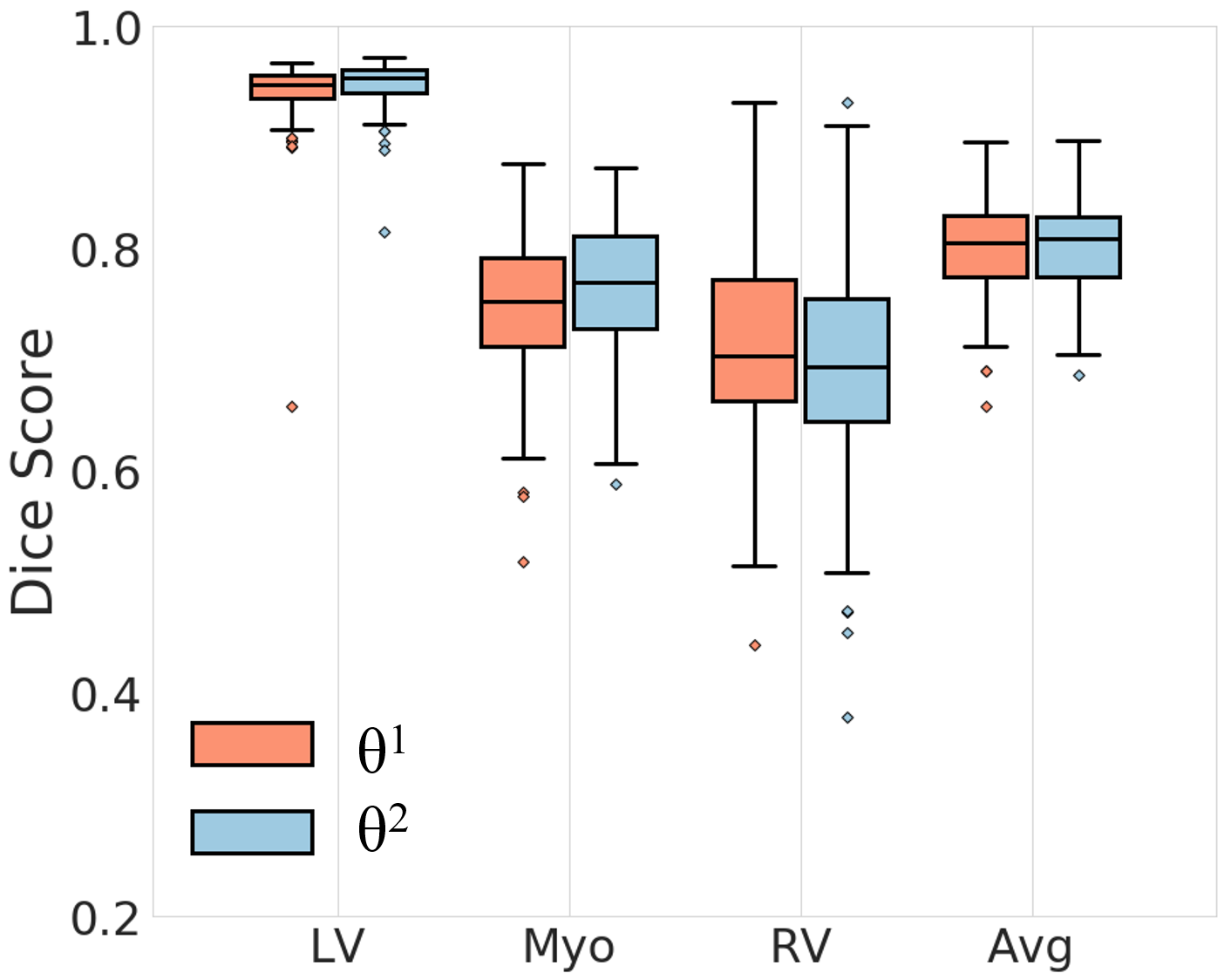}  
  \caption{Two parameterizations}
  \label{fig:sub-weightsharing}
\end{subfigure}
\begin{subfigure}{0.33\textwidth}
  \centering
  \includegraphics[width=0.99\linewidth,height=0.81\textwidth]{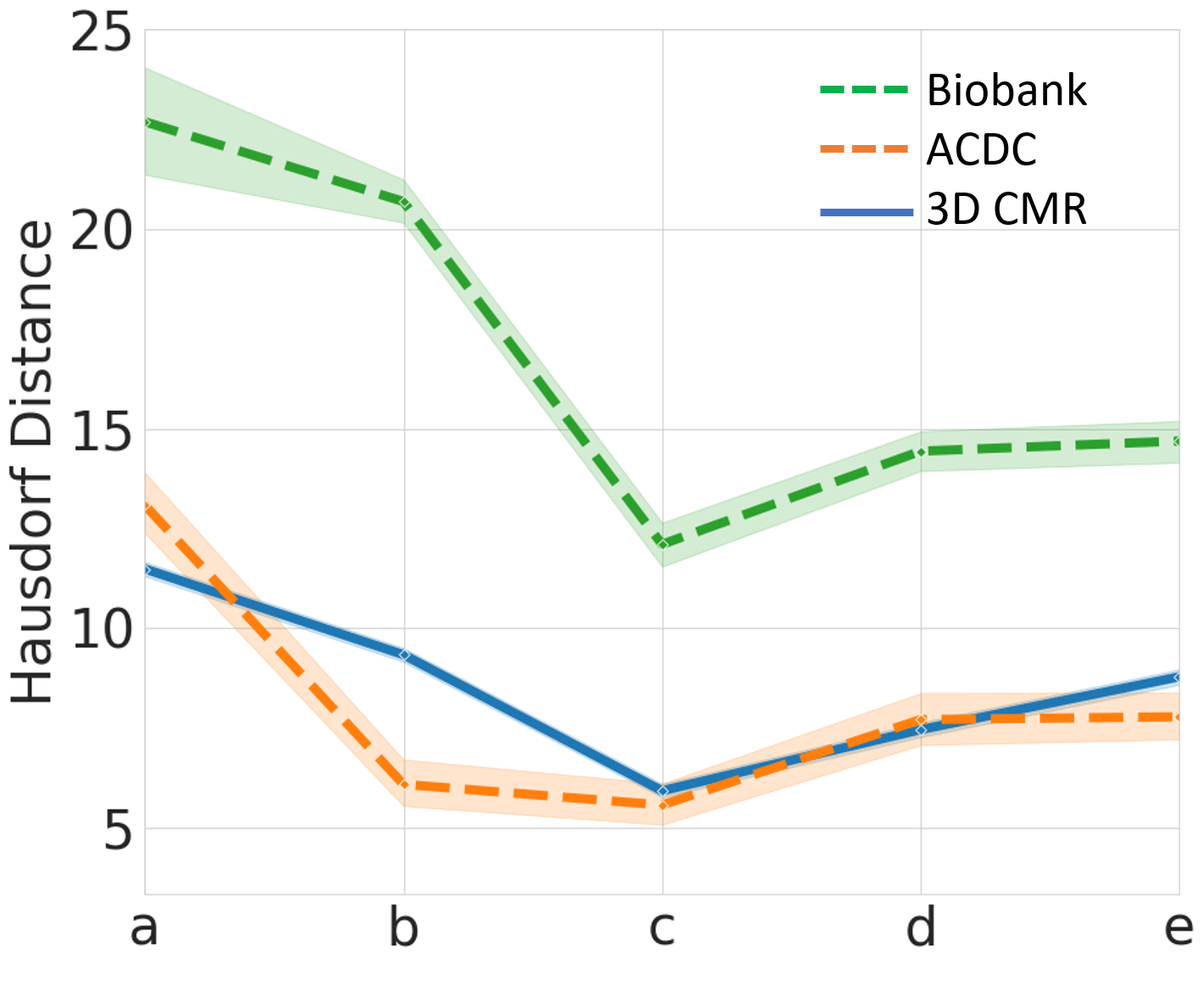}  
  \caption{Impact of using different $\alpha$}
  \label{fig:sub-alpha}
\end{subfigure}
\begin{subfigure}{0.33\textwidth}
  \centering
  \includegraphics[width=0.99\linewidth,height=0.81\textwidth]{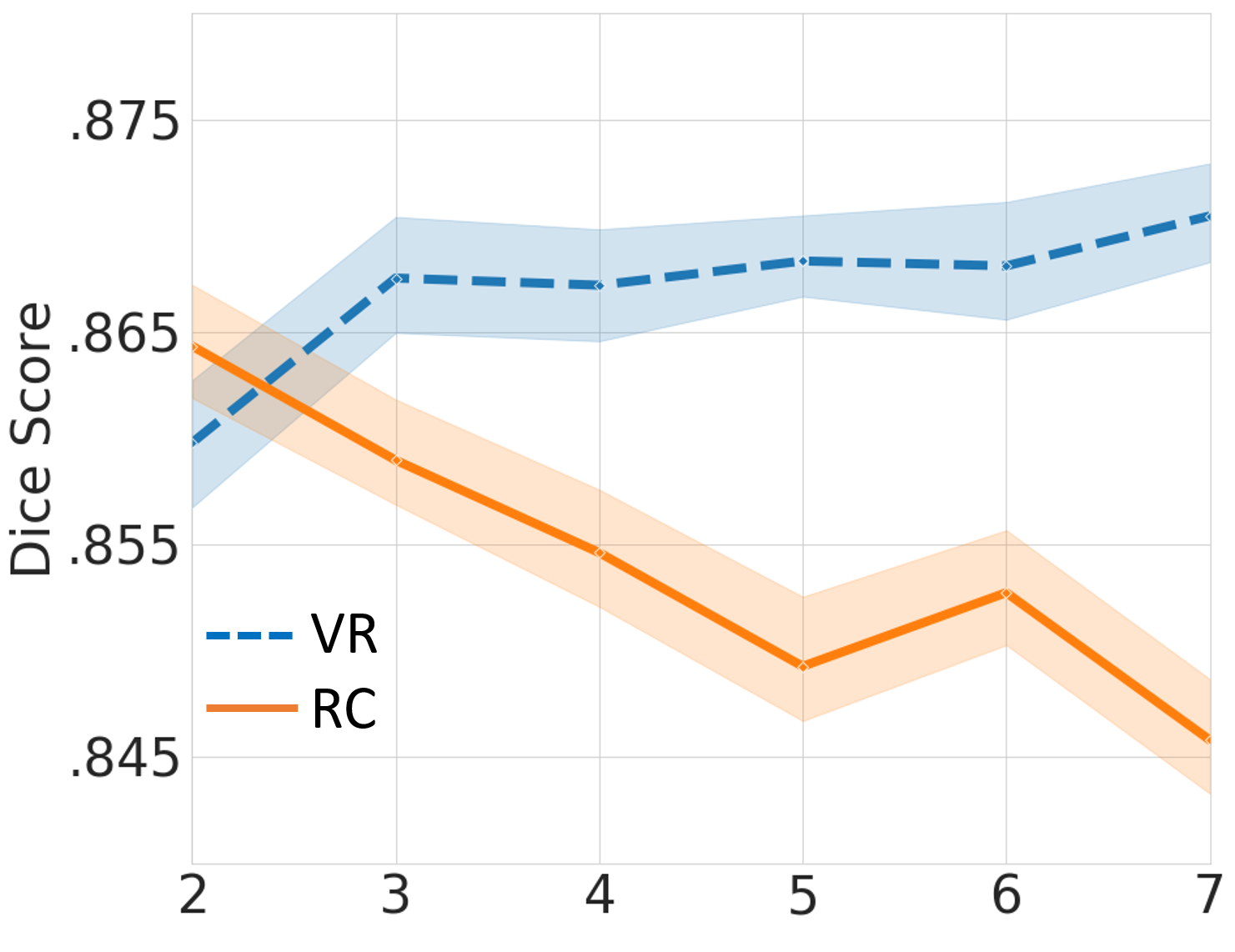}  
  \caption{Cascade Numbers}
  \label{fig:sub-cascadenumber}
\end{subfigure}

\caption{(a): Dice scores of R-$\rm{L}_2$-6$\times$1 using the two parameterizations in Sec.~\ref{sec:para} on the UK Biobank. (b) Impact of using different $\alpha$ in terms of Hausdorff distance on the three datasets. (c): Comparing VR-Net and RC-Net \cite{Zhao_2019_ICCV} using a different number of cascades on the ACDC dataset.}
\label{fig:visualresult}
\end{figure*}

%

\subsection{Ablation Studies}
In this section, we test different configurations for VR-Net. Specifically, we explore the impact of using different data terms, denoising networks,  parameterizations and varying numbers of cascades. For simplicity, we use shorthand notations to represent different configurations. For example, R-$\rm{L}_1$-3$\times$2 indicates that we use the U-Net with residual connection, Eq. \eqref{eq:soft1} ($L_1$ data term), $N_{warp}=3$ and $N_{iter}=2$ in VR-Net. U-$\rm{L}_2$-6$\times$1 indicates that we use the U-Net without residual connection, Eq. \eqref{eq:ShermanMorrison} ($L_2$ data term), $N_{warp}=6$ and $N_{iter}=1$.  

We first compare the results obtained by using different cascades in VR-Net. From Table \ref{tab:cascade}, we observe that the best results almost all come from using 6 cascades (either 3$\times$2 or 6$\times$1), indicating that increasing cascade number improves the performance. On the UK Biobank, the best result is achieved by R-$\rm{L}_2$-6$\times$1 (0.804 Dice and 10.26 HD), while on ACDC the best result is achieved by R-$\rm{L}_1$-3$\times$2 (0.873 Dice and 6.33 HD). When comparing the best performance among different data terms, $L_1$ performs worse than $L_2$ on UK Biobank, while on ACDC $L_1$ is better. This suggests that the proposed VR-Net is robust to different data terms. Next, we compare the results obtained by using different denoising networks, and we notice a tiny improvement when a residual connection is applied.



In Fig.~\ref{fig:sub-weightsharing} we show the performance of VR-Net on UK Biobank with two different parameterizations: ${\bf{\Theta}}^1$ and ${\bf{\Theta}}^2$. From these boxplots, we see that using ${\bf{\Theta}}^2$ performs better on RV and Myo anatomical structures, while on RV using ${\bf{\Theta}}^1$ is better. The averaged results (last two columns) on the three anatomical regions indicate a similar performance between the two parameterizations. Note that the number of network parameters in ${\bf{\Theta}}^1$ is 1/6 of that in ${\bf{\Theta}}^2$.

\reviewertwo{While the original regularization weight $\lambda$ is absorbed in the $\boldsymbol{v}$-subproblem to avoid manual choice, by using the training loss in Eq. \eqref{eq:loss} we do introduce another parameter $\alpha$. However, tuning $\alpha$ is based on the whole dataset and we tune it only during training. We presented a curve plot that illustrates how different $\alpha$ affect the registration accuracy (as shown in Fig.~\ref{fig:sub-alpha}). Specifically, we used five different values of $\alpha$ to train the proposed VR-Net five times on three datasets, i.e., $\alpha_{UKBB}=\{1,0.5,0.1,0.05,0\}$, $\alpha_{ACDC}= \{0.5,0.1,0.05,0.005,0\}$, and  $\alpha_{3D CMR}=\{0.01,0.001,0.0001,0.00001,0\}$. We then plot their registration accuracy (in terms of Hausdorff Distance) on each dataset as $\alpha$ varies. As suggested by the curve plot, the optimal values of $\alpha$ for UK Biobank, ACDC, and 3D CMR datasets are 0.1, 0.05, and 0.0001, respectively.}

\subsection{Initialization Strategies}

\label{sec:initialResult}
\reviewertwo{
In Table \ref{tab:initialization}, we explore the performance of VR-Net using different initialization approaches on the UK Biobank dataset. As is evident in this table, with zeros or noises as the initial displacements, the Dice results of VR-Net dropped by $6.0\%$ and $6.4\%$, respectively, and the HD results dropped by 1.59$mm$ and 1.42$mm$, respectively. These results suggest that making the initialization learnable is crucial as (1) registration is nonconvex and its solution depends on initialization, and (2) our network builds on iterative optimization methods and thus also relies on initialization. Furthermore, our VR-Net is derived using the Taylor linearization and as such computes only a small displacement in each iteration. \reviewerone{When we} initialize the input displacement with noise or zeros, 6 iterations are not sufficient to perform a good registration.}

\begin{table}[h]
    \centering
        \caption{Performance of VR-Net on Biobank using different initialization. \reviewerone{Note the U-Net is not pretrained, it is also a learnable layer in the whole VR-Net.}}
\begin{tabular}{ccc}
\hline
Initialization & Dice       & HD           \\ \hline
U-Net         & .804(.043) & 10.26(3.07) \\ 
Noise         & .740(.048)       &  11.68(3.05)            \\ 
Zeros      & .744(.051)       & 11.85(3.18)             \\ \hline
\end{tabular}

    \label{tab:initialization}
    \vspace{-5pt}
\end{table}

\subsection{Comparison with State-of-the-Art}

\begin{table*}[h]
\centering
\caption{Comparison of image registration performance using different methods on UK Biobank. `Avg' means that Dice (HD) is computed by averaging that of LV, Myo and RV of all subjects in the test set. Here mean and standard deviation (in parenthesis) are reported. 'Unreg' stands for unregistered and \#$\times$RC-Net the number of cascades used in RC-Net. }
\resizebox{0.99\textwidth}{!}{
\begin{tabular}{ccccclcccclcc}
\cline{1-5} \cline{7-10} \cline{12-13} 
\multirow{2}{*}{Methods} & \multicolumn{4}{c}{Dice} &  & \multicolumn{4}{c}{HD} & & \multirow{2}{*}{$J_{ < 0}\%$} & \multirow{2}{*}{$| \nabla J | $ }\\ \cline{2-5} \cline{7-10}
 & LV & Myo & RV & Avg &  & LV & Myo & RV & Avg & &  &\\ \cline{1-5} \cline{7-10} \cline{12-13}
Unreg & .634(.072) & .344(.086) & .551(.080) & .510(.055) &  & 11.99(1.64) & 10.08(2.91) & 24.52(6.24) & 15.53(2.40) & & --  & --\\
FFD & .934(.025) & .711(.081) & .672(.110) & .772(.051) &  & 4.87(2.15) & 7.86(4.03) & 21.18(7.69) & 11.30(3.26)& & 0.23(0.29) & .019(.011)\\
TV-$\rm{L}_1$ & .937(.036) & .717(.076) & .701(.105) & .785(.047) &  & 4.75(1.67) & 7.12(3.16) & 19.73(7.21) & 10.53(2.86)& & 0.65(0.30) & .051(.017) \\ \cline{1-5} \cline{7-10} \cline{12-13}
Siamese & .932(.022) & .706(.069) & .695(.099) & .778(.046) &  & 4.75(1.65) & 6.52(3.23) & 20.69(7.02) & 10.65(3.01) & & 0.42(0.21) & .065(.016)\\
VoxelMorph & .931(.029) & .717(.072) & .685(.102) & .778(.047) &  & 4.57(1.47) & 6.71(3.43) & 21.78(7.27) & 10.69(3.06) & & 0.07(0.10) & .027(.008)\\
2$\times$RC-Net & .942(.022) & .737(.066) & .703(.099) & .794(.044) &  & 4.43(1.57) & 6.65(3.35) & 20.29(7.15) & 10.46(3.01)  & & 0.22(0.17) & .041(.010)\\
3$\times$RC-Net & .944(.036) & .736(.068) & \textbf{.705(.105)} & .795(.048) &  & 4.28(1.81) & 7.39(3.32) & \textbf{19.96(6.53)} & 10.55(2.80)  & & 0.70(0.36) &.069(.019) \\
4$\times$RC-Net & .945(.022) & .736(.065) & .701(.120) & .794(.045) &  & 4.36(1.54) & 7.23(3.39) & 20.54(7.32) & 10.71(2.98) & & 0.49(0.24) & .056(.015) \\
5$\times$RC-Net & .944(.033) & .723(.065) & .703(.109) & .790(.045) &  & 4.24(1.60) & 8.09(3.51) & 20.16(6.83) & 10.83(2.77) & & 1.18(0.53) & .094(.023)  \\
6$\times$RC-Net & .941(.024) & .714(.068) & .696(.114) & .784(.048) &  & 4.60(1.53) & 7.66(3.23) & 20.51(7.17) & 10.92(2.98) & & 1.29(0.55) & .096(.023) \\
7$\times$RC-Net & .943(.025) & .721(.066) & .695(.113) & .786(.047) &  & 4.52(1.56) & 7.66(3.20) & 20.42(7.06) & 10.87(2.89) & & 1.00(0.44) & .084(.022) \\ \cline{1-5} \cline{7-10} \cline{12-13}
R-$\rm{L}_2$-6$\times$1 & \textbf{.948(.021)} & \textbf{.764(.060)} & .700(.105) & \textbf{.804(.043)} && \textbf{3.90(1.41)} & \textbf{6.49(3.79)} & 20.38(7.21) & \textbf{10.26(3.07)} & & 0.38(0.18) & .039(.012)  \\
\cline{1-5} \cline{7-10} \cline{12-13}
\end{tabular}}
\label{tab:ukbb}
\end{table*}

\begin{table*}[h]
\centering
\caption{Comparison of image registration performance using different methods on the ACDC dataset.}
\resizebox{\textwidth}{!}{
\begin{tabular}{ccccclcccclcc}
\cline{1-5} \cline{7-10} \cline{12-13} 
\multirow{2}{*}{Methods} & \multicolumn{4}{c}{Dice} &  & \multicolumn{4}{c}{HD} & & \multirow{2}{*}{$J_{ < 0}\%$} & \multirow{2}{*}{$| \nabla J | $ }\\ \cline{2-5} \cline{7-10}
 & LV & Myo & RV & Avg &  & LV & Myo & RV & Avg & &  &\\ \cline{1-5} \cline{7-10} \cline{12-13}
Unreg & .666(.178) & .540(.143) & .672(.145) & .626(.108) & & 12.21(4.34) & 7.65(2.67) & 12.74(4.56) & 10.87(3.13) & & -- & --\\
FFD & .920(.063) & .792(.067) & .803(.126) & .838(.059) &  & 5.16(2.14) & 5.87(2.18) & 9.60(4.56) & 6.88(2.40) & & 0.32(0.42) & .031(.037)\\
TV-$\rm{L}_1$ & .902(.106) & .793(.086) & .835(.117) & .843(.075) &  & 5.97(3.25) & 6.11(2.85) & 9.51(4.49) & 7.20(2.81) & & 0.52(0.37) & .053(.020)\\ \cline{1-5} \cline{7-10}  \cline{12-13}
Siamese & .872(.106) & .723(.113) & .778(.132) & .791(.081) &  & 7.34(3.49) & 6.05(1.84) & 10.55(4.30) & 7.98(2.68) & & 0.15(0.17) & .052(.011)\\
VoxelMorph & .924(.066) & .789(.096) & .837(.104) & .850(.062) &  & 5.51(2.81) & 5.83(2.26) & 9.33(4.02) & 6.89(2.50) & & 0.38(0.35) & .066(.018) \\
2$\times$RC-Net & .931(.053) & .798(.083) & .864(.082) & .864(.050) &  & 5.46(2.49) & 6.22(2.56) & 8.63(3.97) & 6.77(2.46) & & 0.54(0.37) & .097(.023)\\
3$\times$RC-Net & .931(.048) & .794(.076) & .852(.095) & .859(.049) &  & 6.08(2.61) & 7.02(2.71) & 9.15(3.95) & 7.42(2.40) & & 1.02(0.57) & .131(.029)\\
4$\times$RC-Net & .926(.056) & .789(.075) & .849(.086) & .855(.050) &  & 6.01(2.78) & 6.62(2.70) & 9.32(3.76) & 7.32(2.53) & & 0.95(0.55) & .123(.028)\\
5$\times$RC-Net & .919(.072) & .780(.075) & .849(.091) & .849(.055) &  & 6.38(2.80) & 7.29(2.96) & 9.37(3.78) & 7.68(2.40) & & 1.24(0.68)  &.140(.031)\\
6$\times$RC-Net & .926(.050) & .779(.088) & .853(.090) & .853(.051) &  & 6.67(2.73) & 8.02(3.69) & 9.19(3.89) & 7.96(2.63) & & 1.94(0.91) & .180(.040)\\
7$\times$RC-Net & .927(.048) & .769(.085) & .842(.104) & .846(.053) &  & 6.72(2.57) & 8.44(3.35) & 9.44(3.86) & 8.20(2.50)  & & 2.17(1.00) & .194(.042)\\ \cline{1-5} \cline{7-10}  \cline{12-13}
R-$\rm{L}_1$-3$\times$2 & \textbf{.934(.052)} & \textbf{.815(.078)} & \textbf{.869(.082)} & \textbf{.873(.050)}& & \textbf{5.09(2.20)} & \textbf{5.48(2.19)} & \textbf{8.43(3.72)} & \textbf{6.33(2.13)} & & 0.32(0.25) & .078(.024)\\
\cline{1-5} \cline{7-10} \cline{12-13}
\end{tabular}} \label{tab:acdc}
\end{table*}

\begin{figure*}[t]
\centering  
{\includegraphics[width=0.99\textwidth]{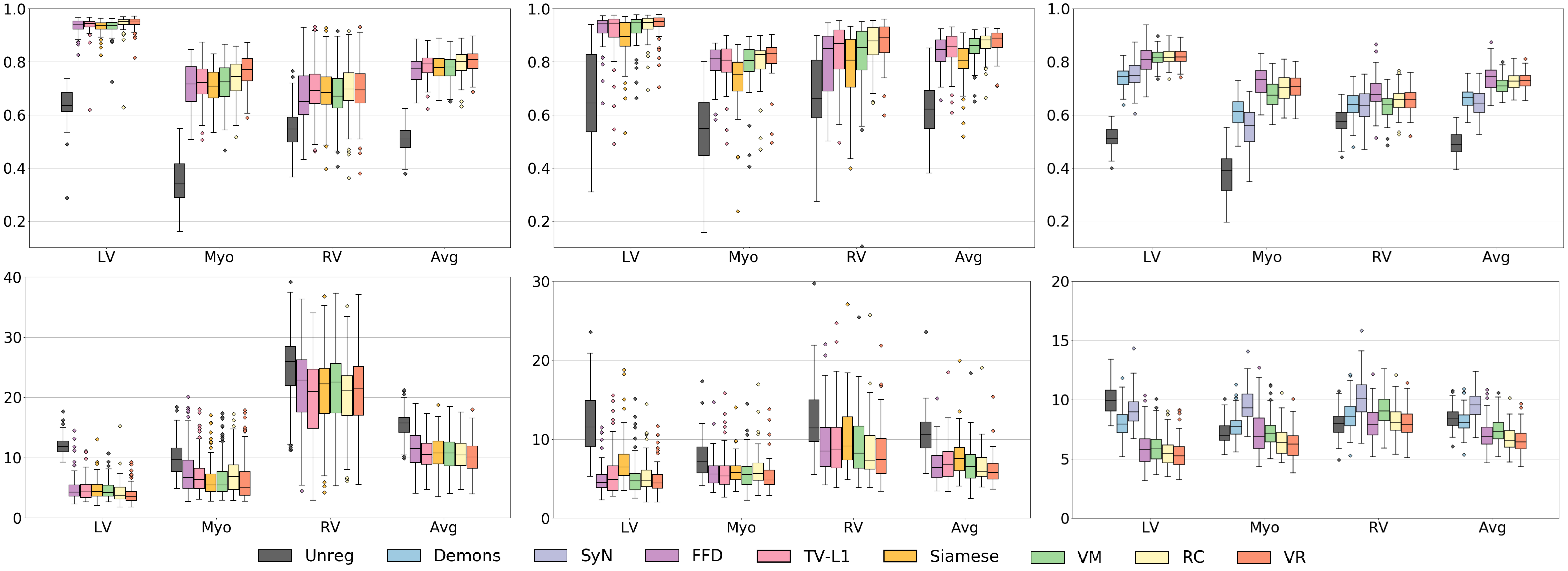}}
\caption{Boxplot illustration of Dice (top row) and HD (bottom row) results obtained by different registration methods on the UK Biobank (left), ACDC (middle), and the 3D CMR (right) datasets. The proposed VR-Net outperforms all compared methods on the UK Biobank and ACDC datasets. Although the Dice of VR-Net is lower than that of FFD on the 3D CMR dataset, it achieves the best HD score. }
\label{fig:boxplots_all}
\end{figure*}

\label{sec:sota}
In this section, we compare our VR-Net with iterative methods (i.e. FFD \cite{rueckert1999nonrigid} and TV-$\rm{L}_1$ \cite{zach2007duality}) 
and data-driven deep learning methods (i.e. VoxelMorph \cite{balakrishnan2018unsupervised,balakrishnan2019voxelmorph}, Siamese network \cite{qin2018joint,qiustacom19} and RC-Net \cite{Zhao_2019_ICCV}) 
on the UK Biobank, ACDC and \reviewerthr{3D CMR dataset}. An overview of the Dice and HD scores of different registration methods can be found in the boxplots in Fig.~\ref{fig:boxplots_all}.

\textbf{2D Methods:} For FFD, we use the implementation in MIRTK \cite{rueckert1999nonrigid}, where we chose the sum of square difference similarity with bending energy regularisation. We use a 3-level multi-resolution scheme and set the spacing of B-spline control points on the highest resolution to $8mm$. For TV-$\rm{L}_1$, which uses the $L_1$ data term and the total variation regularization, we implement its ADMM solver, in which we use a similar three-level multi-scale strategy for the minimization. \reviewerthr{We implement TV-$\rm{L}_1$ using the same variable splitting and therefore its overall iterative structure is very similar to our VR-Net. However, because TV-$\rm{L}_1$ is cheap to iterate, we can set sufficient numbers of inner iterations (associated with variable splitting) and outer iterations (associated with Taylor expansions) to compute the final deformation. In other words, we tune TV-$\rm{L}_1$ to its maximum capability to compete with our method.} The regularization weights in the two methods are tuned to maximize the accuracy performance on test sets. For the data-driven methods, we first compare our VR-Net with VoxelMorph \cite{balakrishnan2019voxelmorph} which we re-implement for 2D registration. We also compare VR-Net with the Siamese network regularized by the approximated Huber loss \cite{qin2018joint,qiustacom19}. Lastly, for the recursive cascade network (RC-Net) \cite{Zhao_2019_ICCV}, which used a 3D U-Net-like architecture in a cascade fashion, we re-implement a 2D version. Overall, the backbone of both VoxelMorph and RC-Net is a U-Net and the loss functions (without segmentation loss) are similar to ours.
\reviewerfur{Note that all the compared data-driven methods (including Siamese, VoxelMorph, and RC-Net) are only trained with the training data and no test-time (instance) optimization is adopted. The hyper-parameters of all data-driven methods are tuned individually according to the validation set for a fair comparison.}


\textbf{3D Methods:} \reviewerthr{We again use a three-level pyramid scheme with SSD similarity and bending energy regularisation for our FFD comparison, tuning control point spacing on the validation set.  We compare with the official ANTs SyN implementation \cite{avants_ANTS} with cross correlation and a three-level pyramid scheme. Cross correlation radius, smoothing $\lambda$, and number of iterations are selected through experimentation on a single validation image rather than the entire set due to the long run time per sample. Also using a three-level pyramid scheme, we finally compare with the SimpleITK \cite{SimpleITK} implementation of the diffeomorphic demons algorithm\cite{vercauteren2009diffeomorphic}, optimizing the number of iterations and smoothing parameter on the validation set.}

\begin{table*}[h]
\centering
\caption{Comparison of image registration performance using different methods on the 3D CMR dataset.}
\resizebox{\textwidth}{!}{
\begin{tabular}{ccccclcccclcc}
\cline{1-5} \cline{7-10} \cline{12-13} 
\multirow{2}{*}{Methods} & \multicolumn{4}{c}{Dice} &  & \multicolumn{4}{c}{HD} & & \multirow{2}{*}{$J_{ < 0}\%$} & \multirow{2}{*}{$| \nabla J | $ }\\ \cline{2-5} \cline{7-10}
 & LV & Myo & RV & Avg &  & LV & Myo & RV & Avg & &  &\\ \cline{1-5} \cline{7-10} \cline{12-13}

Unreg                    & .516(.039) & .384(.084) & .579(.044) & .493(.043) &   & 9.99(1.15) & 7.22(.10) & 8.00(1.11)  & 8.40(.89)&    & --  & --               \\ 
Demons                   & .741(.037) & .607(.056) & .639(.051) & .662(.039) &   & 8.05(1.31) & 7.76(1.08) & 8.67(1.32)  & 8.16(.97)&    & 0.90(0.31)     & .070(.005)   \\ 
FFD                      & .808(.055) & \textbf{.726(.059)} & \textbf{.684(.061)} & \textbf{.739(.047)} &   & 5.89(1.52) & 7.26(1.78) & 7.98(1.38)  & 7.04(1.20)&    & 0.77(0.30)   & .034(.008)      \\ 
SyN                      & .749(.049) & .554(.070) & .629(.064) & .644(.050) &   & 9.08(1.29) & 9.49(1.40) & 10.12(1.60) & 9.56(1.13)&    & 1.46(0.53)  &.045(.008)      \\ \cline{1-5} \cline{7-10} \cline{12-13}
VoxelMorph               & .817(.028) & .676(.051) & .634(.046) & .709(.032) &   & 5.94(1.35) & 7.22(1.26) & 9.13(1.31)  & 7.43(1.05)&    & 0.75(0.22) & .089(.014)      \\ 
2$\times$RC-Net & .820(.030) & .701(.051) & .657(.047) & .726(.034)&  & 5.60(1.25)& 6.50(1.22)  & 8.17(1.21)  & 6.76(0.99)& & 0.11(0.05) & .057(.008)                   \\ 
3$\times$RC-Net   & .824(.027) & .692(.050) & .647(.048) & .721(.033)& & 5.67(1.10) & 6.87(1.24) & 8.61(1.23)  & 7.05(0.96)& & 0.49(0.13) & .085(.013)                  \\ \cline{1-5} \cline{7-10} \cline{12-13}
R-$\rm{L}_2$-2$\times$1 & \textbf{.825(.026)} & .695(.050) & .649(.047)  & .723(.032) & &5.56(1.30) & 6.80(1.23) & 8.61(1.25) & 6.99(1.04) & & 1.07(0.31) & .110(.018)                 \\
R-$\rm{L}_1$-2$\times$1 & .821(.026) & .706(.046) & .650(.046)  & .726(.031) & &5.42(1.25) & 6.57(1.26) & 8.29(1.26) & 6.76(1.03) & & 0.84(0.25)  & .120(.017)                \\
U-$\rm{L}_2$-2$\times$1 & .822(.027) & .674(.051) & .645(.046)  & .714(.033) & &\textbf{5.37(1.16)} & 6.76(1.07) & 8.52(1.29) & 6.89(0.98) & & 0.58(0.25) & .069(.011)                 \\
U-$\rm{L}_1$-2$\times$1 & .821(.028) & .706(.045) & .657(.047)  & .728(.031) & &5.38(1.28) & \textbf{6.23(1.14)} & \textbf{8.03(1.21)} & \textbf{6.55(1.01)} & & 0.24(0.12)    & .081(.011)              \\\cline{1-5} \cline{7-10} \cline{12-13}
\end{tabular}
}
\label{tab:3dresult}
\end{table*}

In Table \ref{tab:ukbb} and \ref{tab:acdc}, we show the quantitative results obtained by using different methods on UK Biobank and ACDC. In the tables, one can see that VR-Net outperforms iterative methods and data-driven methods on both datasets for almost all anatomical structures. On UK Biobank, RC-Net achieves the best result on RV in terms of both Dice and HD of 0.005 and 0.42$mm$ respectively compared to that obtained by our best configuration of (R-$\rm{L}_2$-6$\times$1). However, in terms of Dice, VR-Net achieves 0.948 on LV and 0.764 on Myo, outperforming $3\times$RC-Net by 0.004 and by 0.028, respectively. In terms of HD for LV and Myo, our VR-Net improves $3\times$RC-Net from 4.28$mm$ to 3.90$mm$ and from 7.39$mm$ to 6.49$mm$, respectively. On average, the proposed VR-Net achieves a better Dice and HD score than $3\times$RC-Net, making our VR-Net the best method on this dataset.

\begin{figure*}[ht]
\begin{subfigure}{0.5\textwidth}
  \centering
  \includegraphics[width=0.99\linewidth]{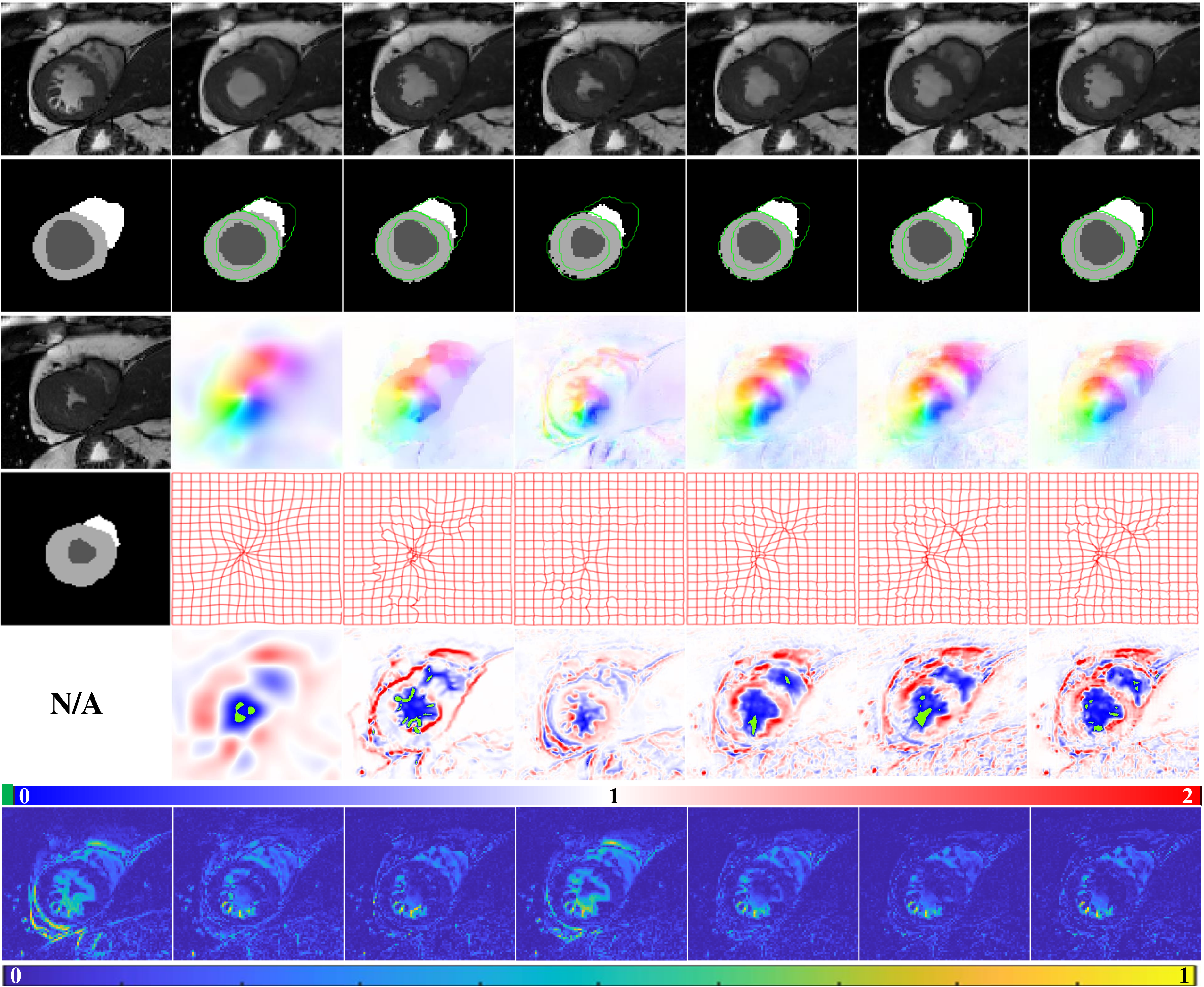}  
  \caption{2D ACDC}
  \label{fig:sub-first}
\end{subfigure}
\begin{subfigure}{0.5\textwidth}
  \centering
  \includegraphics[width=0.99\linewidth]{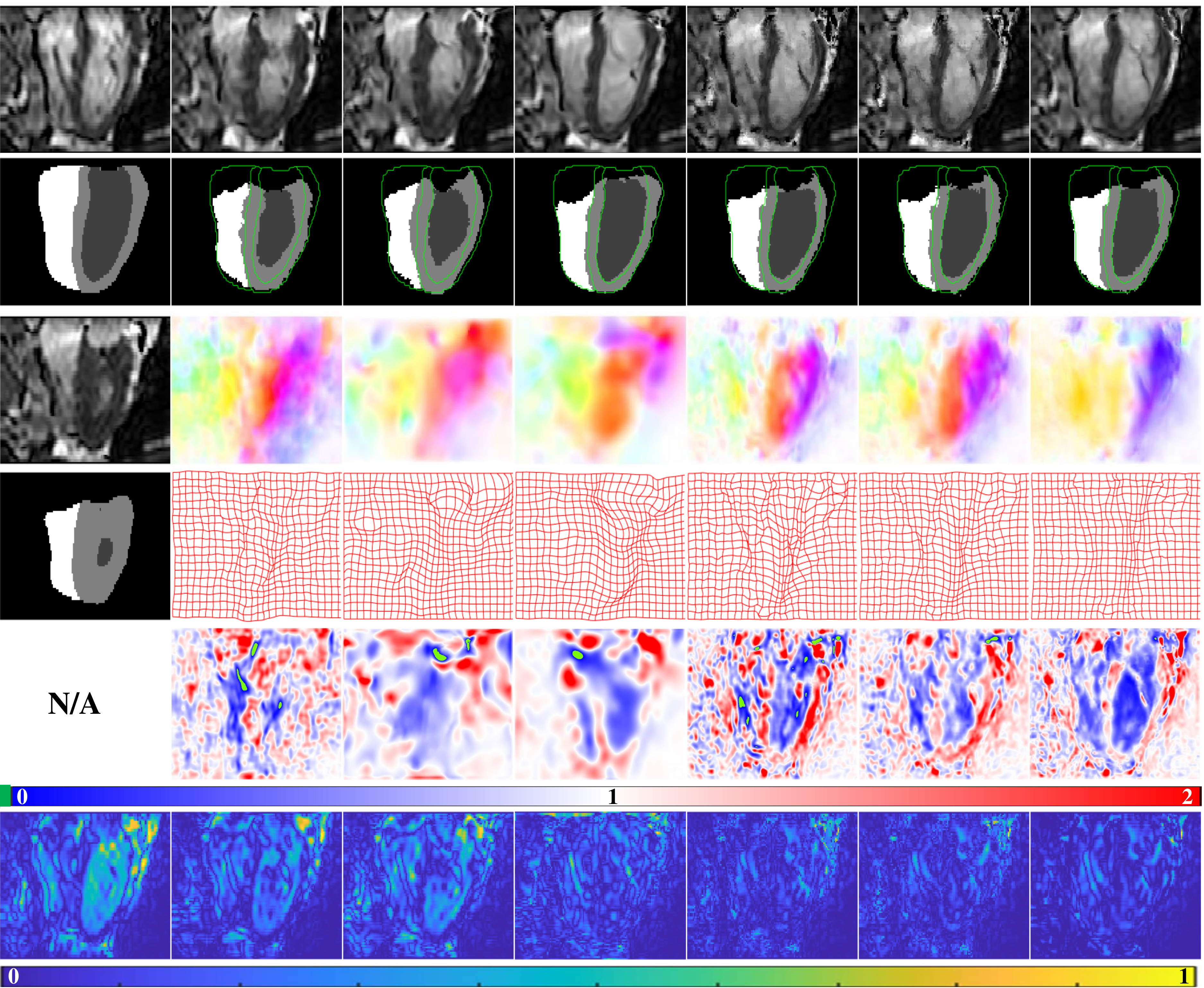}  
  \caption{3D CMR}
  \label{fig:sub-second}
\end{subfigure}
\caption{Comparing visual results obtained by different registration methods on the ACDC and 3D CMR datasets. The 1st column includes ED image, ED mask, ES image, ES mask, and absolute difference between ES image and the ED image. Excluding the 1st column, for (a) ACDC, from left to right: FFD, TV-$\rm{L}_1$, Siamese Net, VoxelMorph, RC-Net, and VR-Net results, respectively, for (b) 3D CMR, from left to right: Demons, ANTs SyN, FFD, VoxelMorph, RC-Net, and VR-Net results, respectively. From top to bottom: warped ES images, warped ES masks (with ground truth mask shown in green contours), estimated deformations (shown in HSV and grid), the Jacobian map, and absolute differences between warped ES images and the ground truth ED image, respectively.}
\label{fig:visualresult}
\end{figure*}

On ACDC, the proposed VR-Net with the configuration of R-$\rm{L}_1$-3$\times$2 outperforms all other methods across all anatomical structures. While $2\times$RC-Net also obtains comparable results, one can notice that its performance drops rapidly with more cascades. To visualize this, we plotted the average Dice scores of both RC-Net and VR-Net versus the number of cascades in Fig.~\ref{fig:sub-cascadenumber} on this dataset. As is evident from this figure, there is a sharp decrease in the performance of RC-Net, which is due to RC-Net overfitting the small training set of 40 subjects. In contrast, VR-Net performs constantly well using an increasing number of cascades, demonstrating its data-efficiency. This is attributable to the integration of the iterative variational model(prior knowledge) into the VR-Net.

\reviewerthr{On the 3D CMR dataset, as listed in Table \ref{tab:3dresult}, FFD outperforms all compared methods on the Myo and RV, and achieves the highest average Dice score, i.e. 0.739. Although the average Dice of our VR-Net (U-$\rm{L}_1$-2$\times$1) is lower than that of FFD with 0.01 margin, the average HD score is higher than that of FFD. Furthermore, the proposed VR-Net achieves both the highest Dice and HD score among the compared data-driven methods.}

\reviewerone{We also listed the percentage of negative Jacobian determinant values as well as the gradient magnitude of the Jacobian determinant of all compared methods on both the UK Biobank and ACDC datasets. From Table \ref{tab:ukbb} and \ref{tab:acdc}, we can see that although VR-Net generates foldings in deformation, it produces fewer than RC-Net with the same number of cascades, i.e. $0.38\%$ of R-$\rm{L}_2$-6$\times$1 and $1.00\%$ of 7$\times$RC-Net on the UK Biobank,  and $0.32\%$ of R-$\rm{L}_1$-3$\times$2 and $2.17\%$ of 7$\times$RC-Net on ACDC.
 On the 3D CMR dataset, as shown in Table \ref{tab:3dresult}, VR-Net ($0.24\%$) again outperforms the 3$\times$RC-Net ($0.49\%$) as well as VoxelMorph ($0.75\%$), however, it is lower than the 2$\times$RC-Net ($0.11\%$). Overall, VR-Net cannot guarantee zero foldings in estimated deformations, it produces deformations comparable with VoxelMorph and RC-Net.
}

In Table \ref{tab:runtime}, we list the runtime of different methods. Although we adopt the mathematical structure of a variational model, our VR-Net is very close to the purely data-driven deep learning methods as the solutions are point-wise closed-form, and it is much faster than traditional iterative methods. The runtime is measured and averaged over 100 test subjects. 

Lastly, in Fig.~\ref{fig:visualresult}, we compare the visual results of different methods by showing two image registration examples from the ACDC and 3D CMR datasets. On the ACDC, as can be seen, FFD (2nd column), which used $L_2$ regularization, over-smooths the displacement, the warped ES images are also over-smoothed around the Myo/LV area resulting in the high absolute differences. In contrast, TV-$\rm{L}_1$ (3rd column), which used $L_1$ regularization, preserves edges in the resulting displacements. However, the shape of RV warped by TV-$\rm{L}_1$ is not very smooth. This side effect also can be seen in the Siamese network result, and the Siamese network also produces a very high difference map. The displacement results of VoxelMorph, RC-Net, and VR-Net are smooth and look more natural. But the absolute difference map of VoxelMorph shows the less accurate registration than VR-Net. Additionally, the Jacobian map of RC-Net has more foldings than VR-Net (highlighted in green). The warped Myo of VoxelMorph from ACDC has unsmooth shape. The unsmooth shape can also be found in the warped masks of RC-Net. In terms of similarity, the result of VR-Net is the closest one to the ground truth, visually illustrating that the method is more accurate for image warping.

On the 3D CMR, Demons and ANTs SyN produce very high absolute difference maps and the warped ES masks are less accurate than the other compared methods. The warped ES mask of FFD has a very good overlapping with the ground truth ED mask. However, its displacement has much foldings (shown in red grid). The foldings of displacements can also be seen in the VoxemMorph, RC-Net, and VR-Net, however, the warped ES image of VR-Net is more close to the ground truth ED image. And the warped ES images of both VoxelMorph and RC-Net have distorted regions on the upper right, resulting in a high difference map on this area.
\begin{table}[h!]
\centering
\caption{Runtimes of different methods. The runtimes are measured and 
averaged over 100 test subjects.}
\resizebox{0.42\textwidth}{!}{
\begin{tabular}{ccccc}
\hline
\multirow{2}{*}{Methods} & \multicolumn{2}{c}{2D} & \multicolumn{2}{c}{3D} \\ \cline{2-5} 
                         & CPU         & GPU       & CPU          & GPU      \\ \hline
TV-$\rm{L}_1$                 & 10.01       & --        & --           & --       \\ 
Demons                   & --          & --        & 11.15         & --       \\ 
SyN                      & --          & --        & 1657.35      & --       \\ 
FFD                      & 5.15        & --        & 141.38         & --       \\ \hline
Siamese                  & 0.07        & 0.01      & --           & --       \\ 
VoxelMorph               & 0.07        & 0.01      & 5.97         & 0.10     \\ 
2$\times$RC-Net          & 0.13        & 0.01      & 11.95        & 0.21     \\ 
3$\times$RC-Net          & 0.19        & 0.02      & 17.85        & 0.34     \\ 
7$\times$RC-Net          & 0.44        & 0.03      & --           & --       \\ \hline
{R,U}-$\rm{L}_{1,2}$-1$\times$1     & 0.13        & 0.01      & 11.95           & 0.22     \\ 
{R,U}-$\rm{L}_{1,2}$-2$\times$1     & 0.19        & 0.02      & 18.25        & 0.33     \\ 
{R,U}-$\rm{L}_{1,2}$-6$\times$1     & 0.42        & 0.03      & --           & --       \\ \hline
\end{tabular}
}
\centering

\label{tab:runtime}
\end{table}

\subsection{Discussion}

\subsubsection{Relationship with VoxelMorph and RC-Net}

\reviewerone{In the proposed VR-Net, we use an additional U-Net to learn initial displacements. We emphasize here that this U-Net is not pre-trained and instead it is part of the VR-Net, which is trained end-to-end. In this case, without any DL, WL, or ICL layers this initial U-Net alone is essentially VoxelMorph, the performance of which is inferior to our VR-Net by a large margin as shown in Tables \ref{tab:ukbb}, \ref{tab:acdc} and \ref{tab:3dresult}. If we recursively use the U-Net for multiple times without using other subsequent layers (such as ICL/GDL), then the model is equivalent to the RC-Net, the performance of which is worse than our VR-Net as shown in Table \ref{tab:ukbb}, \ref{tab:acdc} and \ref{tab:3dresult}.}

\subsubsection{Generalized Denoising Layers}
\begin{figure}[h!]
\vspace{-5pt}
\centering  
{\includegraphics[width=0.49\textwidth]{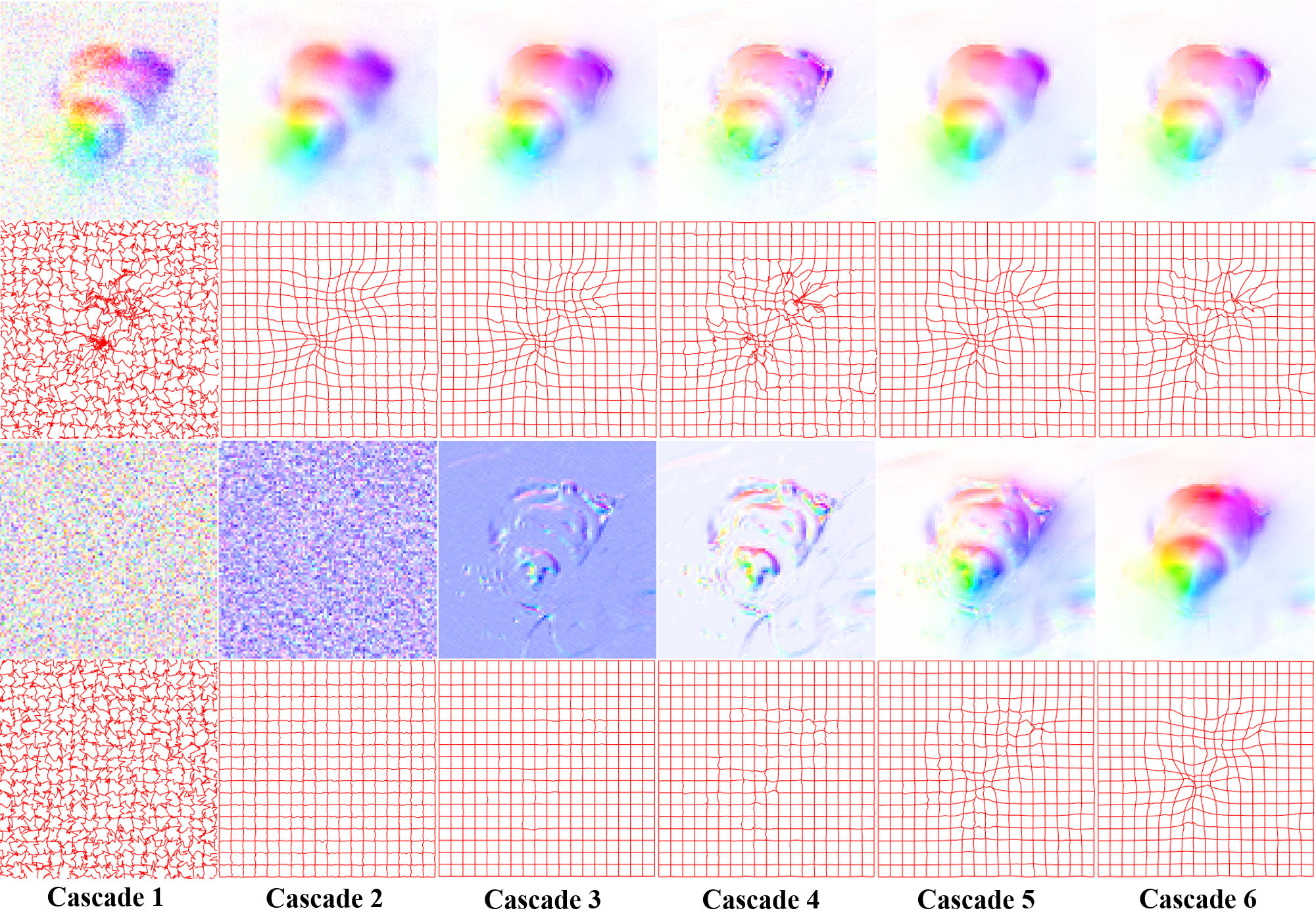}}
\caption{\reviewerone{Visualization of the deformation after GDL in each cascade using noise corrupted deformation (top) and random noise (bottom) as initializations. The top two rows show a noise corrupted deformation is denoised by the VR-Net as cascades proceed. The bottom two rows show if we initialize the input with random noise, VR-Net is still capable of producing a smooth deformation.}}
\vspace{-5pt}
\label{fig:smooth}
\end{figure}

\reviewerone{To understand how the GDL layer is functioning within the network, in Fig.~\ref{fig:smooth} we illustrate the output of this layer after each cascade of the VR-Net using two different setups. Specifically, we use R-$\rm{L}_2$-6$\times1$ using both the U-Net and random noise initialization from Table \ref{tab:initialization}. For the U-Net initialization, we add a Gaussian noise to the input displacement to demonstrate whether this layer can produce any smoothing effect. As shown in the top two rows of Fig.~\ref{fig:smooth}, the deformation becomes gradually smooth as cascades proceed. The deformation also gets increasingly smooth for the random noise initialization. This visualization suggests that our GDL has the denoising effect.}
\reviewerone{However, the capability of GDL is beyond denoising alone. As shown in the last two rows in Fig.~\ref{fig:smooth}, this layer can turn a pure random noise into deformation, indicating its capability of inducing smoothness whilst going beyond denoising and contributing to the deformation itself.}

\subsubsection{Identifying the Optimal Structure} \reviewertwo{In Table \ref{tab:cascade}, we list 24 configurations of VR-Net, along with proposed two parameterizations of ${\bf{\Theta}^1}$ and ${\bf{\Theta}^2}$. Empirically searching for the best structure can be computationally expensive. What is the strategy to efficiently determine the combination? We observe that the best results almost all come from VR-Net with 6 cascades (maximum we can afford) in 2D datasets, indicating that increasing the cascade number improves performance. We therefore suggest using more cascades if one can afford them. As for the two parameterizations ($\Theta^1$ and $\Theta^2$), we notice a slight improvement using $\Theta^2$ and therefore use this parameterization for all our comparative experiments. Comparing different data terms, we find $\rm{L}_1$ and $\rm{L}_2$ are on par with each other, which may be due to their closed-form solutions. We therefore use both $\rm{L}_1$ and $\rm{L}_2$ data terms for comparative experiments. Next, by comparing the results obtained by using different denoising networks in Table \ref{tab:cascade}, we find residual U-Net performs better and therefore use it for the comparative experiments on UK Biobank and ACDC. However, U-Net is better on the 3D CMR dataset, as shown in Table \ref{tab:3dresult}.}

\subsubsection{Brightness constancy assumption}
\reviewerfur{The brightness constancy assumption in Eq. \eqref{eq:variational} is often not suited for medical images with contrast variances and therefore our method will not work well for those images. However, we would like to point out that the proposed framework is not limited to only this assumption and can be extended to other similarity/dissimilarity metrics such as local cross correlation (invariant to multiplicative illumination changes), mutual information (suitable for multi-modality image registration) and others. The idea is to use the second-order Taylor theorem \cite{vogel2013evaluation,werlberger2010motion} to expand a respective metric and then approximate the Hessian matrix in the Taylor expansion with a positive semi-definite matrix. In this case, the resultant problem is a convex optimization which fits in our proposed framework.}

\section{Conclusion}
In this paper, we propose a model-driven VR-Net for deformable image registration, which combines the iterative variational method with modern data-driven deep learning methods. By taking advantage of both approaches, our VR-Net outperforms deep data-driven methods as well as classical iterative methods (in terms of Hausdorff distance) on three cardiac MRI datasets. Extensive experimental results show our VR-Net is fast, accurate, and data-efficient. For our future work, we will extend the VR-Net to multi-modality image registration.

\hypertarget{A1}{}
\section{Appendix 1}
\label{sec:app1}
In this section, we propose to derive the solution of ${\boldsymbol{u}}-${{subproblem}} ($s=1$) in Section~\ref{sec:VS} using a primal-dual method, originally proposed in \cite{chambolle2004algorithm} for Total Variation denoising \cite{rudin1992nonlinear}. Here we use all notations in 3D only. First, we rewrite the subproblem into its discrete form 
\begin{equation} \label{eq:usubproblem}
\mathop {\min }\limits_{ {\boldsymbol{u}} } {\left\| {\rho ({\boldsymbol{u}})} \right\|^1} + \frac{\theta}{2} \|{\boldsymbol{v}}^k - {\boldsymbol{u}}  \|^2,
\end{equation}
where $\rho({\boldsymbol{u}})=  \langle \nabla I_1, {\boldsymbol{u}} -  \boldsymbol{u}^\omega \rangle + I_1 - I_0 $. This minimization problem \eqref{eq:usubproblem} can be converted equivalently to a saddle-point problem by writing the first term as a maximization, i.e.
\begin{equation} \nonumber
     {\left\| {\rho ({\boldsymbol{u}})} \right\|^1}  = \mathop {\max }\limits_{{{\left\| z \right\|}_\infty } \le 1} \mathop {\left\langle { {\rho ({\boldsymbol{u}})}, z} \right\rangle },  
\end{equation}
over the dual variable $z \in {\mathbb{R}{^{MNH}}}$ where $MNH$ is the
the image size, and 
$\left\| z \right\|_{\infty} = {\max }_{i,j,l} \left| {{{z}_{i,j,l}}} \right|$ and 
$\langle  \rho ({\boldsymbol{u}}), z \rangle = \sum_{i,j,l} (\rho ({\boldsymbol{u}}))_{i,j,l} z_{i,j,l}$ 
where $i,j,l$ denote image indices.

The minimization problem \eqref{eq:usubproblem} is equivalent to the following primal-dual (min-max) problem, i.e.
\begin{equation} \label{eq:usubproblempd}
\mathop {\min }\limits_{ {{\boldsymbol{u}}} } \mathop {\max }\limits_{{{z, \left\| z \right\|}_\infty } \le 1} \mathop {\left\langle { {\rho({{\boldsymbol{u}}})}, z} \right\rangle } + \frac{\theta}{2} \|{\boldsymbol{v}}^k - {\boldsymbol{u}}   \|^2,
\end{equation}
over the primal variable ${\boldsymbol{u}} \in {\mathbb{R}{^{MNH}}} $ and the dual variable $z$, respectively. 

First, we differentiate \eqref{eq:usubproblempd} with respect to ${\boldsymbol{u}}$ and derive its first-order optimality condition, resulting in the following closed-form solution for ${\boldsymbol{u}}$
\begin{equation}\label{eq:usolution}
{\boldsymbol{u}} = {\boldsymbol{v}}^k  - z \frac{\nabla I_1}{\theta}.
\end{equation}
We then plug the solution \eqref{eq:usolution} into \eqref{eq:usubproblempd}, converting the primal-dual problem into the following dual problem only
\begin{equation} \label{eq:psubproblempd}
 \mathop {\max }\limits_{{{z, \left\| z \right\|}_\infty } \le 1} \mathop {\langle { {\rho ({{\boldsymbol{v}}^k  - z \frac{\nabla I_1}{\theta}})}, z} \rangle } + \frac{1}{2\theta} \| z {\nabla I_1}\|^2_2.
\end{equation}
If we differentiate \eqref{eq:psubproblempd} with respect to $z$ and derive its first-order optimality condition, we have the following formulation
\begin{equation} \nonumber
\hat{z} = \frac{\theta{\rho ({{\boldsymbol{v}}^k  })}}{|\nabla I_1|^2},
\end{equation}
which needs to be projected to the convex set ${Z} = \left\{ z \in {{{\mathbb{R}^{MNH}}}} : {{{\left\| z \right\|}_\infty } \le 1} \right\}$ to satisfy the constraint ${{{\left\| z \right\|}_\infty } \le 1} $. This results in
\begin{equation}\label{eq:zsolution}
z = \frac{{{{\hat{z} }_{i,j,l}}}}{{\max \left( {\left| {{{\hat{z} }_{i,j,l}}} \right|,1} \right)}}.
\end{equation}

Finally, we plug \eqref{eq:zsolution} into \eqref{eq:usolution} which leads to the solution for ${\boldsymbol{u}}$ without involving the dual variable $z$
\begin{equation}\label{eq:usolution1}
{\boldsymbol{u}} = {\boldsymbol{v}}^k  - \frac{{{{\hat{z} }_{i,j,l}}}}{{\max \left( {\left| {{{\hat{z} }_{i,j,l}}} \right|,1} \right)}} \frac{\nabla I_1}{\theta},
\end{equation}
which is a point-wise, closed-form solution, the same as \eqref{eq:soft1} of the ${\boldsymbol{u}}-${{subproblem}} in Section~\ref{sec:VS}. We highlight that our derivation presented here can be easily applied to vector images, which are usually appearing in data terms that use image patch or gradient information.

\hypertarget{A2}{}
\section{Appendix 2}
In this section, we derive the solution of the Sherman Morrison formula \eqref{eq:ShermanMorrison} in 2D and 3D. For both cases, we need to invert the left-hand side matrix in Eq. \eqref{eq:ShermanMorrison}. As per \cite{bartlett1951inverse}, we have 
\begin{equation} \nonumber
      ({\bf{J}} {\bf{J}} ^{\rm{T}} + \theta {\mathds{1}})^{-1} = \theta^{-1} {\mathds{1}} - \frac{ {\bf{J}} {\bf{J}} ^{\rm{T}} }{\theta^2 + \theta{\bf{J}}^{\rm{T}} {\bf{J}}  }.
\end{equation}
In 2D, this matrix is a $2 \times 2$ symmetric matrix for which each entry is of the 2D image size ($MN$). In 3D, it becomes a $3 \times 3$ symmetric matrix for which each entry is of the 3D image size ($MNH$). The solution $\boldsymbol{u}^{k+1}$ is therefore given by
\begin{equation} \label{eq:matrixvector}
 \boldsymbol{u}^{k+1} =  \boldsymbol{u}^\omega + \left[ {\mathds{1}} - \frac{ {\bf{J}} {\bf{J}} ^{\rm{T}} }{\theta + {\bf{J}}^{\rm{T}} {\bf{J}}  } \right] \left[ {\boldsymbol{v}}^k-\boldsymbol{u}^\omega - \theta^{-1}{\bf{J}}(I_1 - I_0)\right],
\end{equation}
which is the form in terms of matrix and vector multiplication. With Eq.~\eqref{eq:matrixvector}, it is now trivial to derive the final point-wise, closed-form solutions in both 2D and 3D. 

First, in 2D where $I_1 \in  {\mathbb{R}{^{MN}}} $, we have
\begin{equation} \nonumber
    {\bf{J}} {\bf{J}} ^{\rm{T}} = \left[ \begin{array}{l}
{I_1^xI_1^x}\;\;{I_1^xI_1^y}\\
{I_1^yI_1^x}\;\;{I_1^yI_1^y}
\end{array} \right] \in   {\left(\mathbb{R}^{MN}\right)^4}
\end{equation}
and ${\bf{J}}^{\rm{T}} {\bf{J}} = |\nabla I_1|^2 = {I_1^xI_1^x} + {I_1^yI_1^y}$, where $I_1^x \in   {\mathbb{R}^{MN}}$ and $I_1^y \in   {\mathbb{R}^{MN}} $ are respectively the horizontal and vertical derivatives of the source image $I_1$. With $\boldsymbol{u}=(u_1, u_2)^{\rm{T}} \in   {\left(\mathbb{R}^{MN}\right)^2} $ and $\boldsymbol{v}=(v_1, v_2)^{\rm{T}} \in   {\left(\mathbb{R}^{MN}\right)^2} $, we can rewrite Eq.~\eqref{eq:matrixvector} into the following forms in terms of both components of $\boldsymbol{u}$
\begin{equation}\label{eq:s1}
\begin{cases}
{u}_x^{k+1} = {u}_x^\omega + \frac{\begin{array}{c} \displaystyle (I_1^yI_1^y + \theta)({v}_x^{k}-{u}_x^\omega) - I_1^xI_1^y({v}_y^{k}-{u}_y^\omega) \vspace{-3pt} \\ - I_1^x (I_1-I_0) \end{array}}{\displaystyle {I_1^xI_1^x} + {I_1^yI_1^y} + \theta} \vspace{3pt}\\
{u}_y^{k+1} = {u}_y^\omega + \frac{\begin{array}{c} \displaystyle (I_1^xI_1^x + \theta)(\displaystyle {v}_y^{k}-{u}_y^\omega) - I_1^yI_1^x({v}_x^{k}-{u}_x^\omega) \vspace{-3pt}\\ - I_1^y (I_1-I_0) \end{array}}{\displaystyle {I_1^xI_1^x} + {I_1^yI_1^y} + \theta}
\end{cases}.
\end{equation}

Then, in 3D where $I_1 \in  {\mathbb{R}{^{MNH}}} $, we have
\begin{equation} \nonumber
    {\bf{J}} {\bf{J}} ^{\rm{T}} = \left[ \begin{array}{l}
{I_1^xI_1^x}\;\;{I_1^xI_1^y} \;\; I_1^xI_1^z\\
{I_1^yI_1^x}\;\;{I_1^yI_1^y} \;\;{I_1^yI_1^z}\\
{I_1^zI_1^x}\;\;{I_1^zI_1^y} \;\;{I_1^zI_1^z}
\end{array} \right] \in   {\left(\mathbb{R}^{MNH}\right)^9}
\end{equation}
and ${\bf{J}}^{\rm{T}} {\bf{J}} = |\nabla I_1|^2 = {I_1^xI_1^x} + {I_1^yI_1^y} + {I_1^zI_1^z} $, where $I_1^x \in   {\mathbb{R}^{MNH}} $, $I_1^y \in   {\mathbb{R}^{MNH}} $ and $I_1^z \in   {\mathbb{R}^{MNH}} $ are the derivatives of $I_1$ along $x$, $y$ and $z$ directions, respectively. With $\boldsymbol{u}=(u_1, u_2, u_3)^{\rm{T}} \in   {\left(\mathbb{R}^{MNH}\right)^3}$ and $\boldsymbol{v}=(v_1, v_2, v_3)^{\rm{T}} \in   {\left(\mathbb{R}^{MNH}\right)^3} $, we can rewrite Eq.~\eqref{eq:matrixvector} into the following forms in terms of each component of $\boldsymbol{u}$:

\begin{equation} 
\label{eq:s2}
\resizebox{\columnwidth}{!}{$
\begin{cases}
    {u}_x^{k+1} = {u}_x^\omega + \frac{\begin{array}{c}(I_1^yI_1^y + I_1^zI_1^z + \theta)({v}_x^{k} - {u}_x^\omega)  - I_1^xI_1^y({v}_y^{k}  - {u}_y^\omega) \vspace{-3pt}\\ - I_1^xI_1^z({v}_z^{k} - {u}_z^\omega) - I_1^x (I_1-I_0)\end{array}}{\displaystyle {I_1^xI_1^x} + {I_1^yI_1^y} + {I_1^zI_1^z}+ \theta} \vspace{3pt}\\
    {u}_y^{k+1} = {u}_y^\omega + \frac{\begin{array}{c}(I_1^xI_1^x + I_1^zI_1^z + \theta)({v}_y^{k} - {u}_y^\omega) - I_1^yI_1^x({v}_x^{k} - {u}_x^\omega) \vspace{-3pt} \\ - I_1^yI_1^z({v}_z^{k} - {u}_z^\omega) - I_1^y (I_1-I_0) \end{array}}{\displaystyle {I_1^xI_1^x} + {I_1^yI_1^y} + {I_1^zI_1^z}+ \theta} \vspace{3pt}\\
    {u}_z^{k+1} = {u}_z^\omega + \frac{\begin{array}{c}(I_1^xI_1^x + I_1^yI_1^y + \theta)({v}_y^{k} - {u}_y^\omega) - I_1^zI_1^x({v}_x^{k} - {u}_x^\omega) \vspace{-3pt} \\- I_1^zI_1^y({v}_z^{k} - {u}_z^\omega) - I_1^z (I_1-I_0) \end{array}}{\displaystyle {I_1^xI_1^x} + {I_1^yI_1^y} + {I_1^zI_1^z} + \theta}
\end{cases}$
}
\end{equation}
We note that both 2D and 3D solutions, i.e., Eqs. \eqref{eq:s1} and \eqref{eq:s2} are closed-form and point-wise and therefore can be computed very efficiently.

\bibliographystyle{IEEEtran}
\bibliography{IEEEabrv,ref}

\end{document}